\def\paperID{13271}
\def\confName{ICCV}
\def\confYear{2025}

\def\paperTitle{Distilling Parallel Gradients for Fast ODE Solvers of Diffusion Models}

\def\authorBlock{
    Beier Zhu$^1$\thanks{~Equal contribution. $\dagger$ Corresponding author. This work was partially done during Beier Zhu's visit at WestLake University.} \qquad
    Ruoyu Wang$^2$\footnotemark[1] \qquad
    Tong Zhao$^2$ \qquad Hanwang Zhang$^1$ \qquad Chi Zhang$^{2\dagger}$\\
    $^1$Nanyang Technological University, $^2$Westlake University \\
    {\tt\small \{beier.zhu, hanwangzhang\}@ntu.edu.sg, \{wangruoyu71, zhaotong68, chizhang\}@westlake.edu.cn}
}

\newif\ifreview 
\newif\ifarxiv 
\newif\ifcamera \newcommand{\cameraready}{\cameratrue}
\newif\ifrebuttal 

\cameraready 

\pdfoutput=1
\documentclass[10pt,twocolumn,letterpaper]{article}
\ifreview \usepackage[review]{cvpr} \fi
\ifarxiv \usepackage[pagenumbers]{cvpr} \fi
\ifrebuttal \usepackage[rebuttal]{cvpr} \fi
\ifcamera \usepackage{cvpr} \fi

\usepackage{amsfonts,bm}

\usepackage{graphicx}	
\usepackage{amsmath}	
\usepackage{amssymb}	
\usepackage{booktabs}
\usepackage{times}
\usepackage{microtype}
\usepackage{epsfig}
\usepackage{caption}
\usepackage{float}
\usepackage{placeins}
\usepackage{color, colortbl}
\usepackage{stfloats}
\usepackage{enumitem}
\usepackage{tabularx}
\usepackage{xstring}
\usepackage{multirow}
\usepackage{xspace}
\usepackage{url}
\usepackage{subcaption}
\usepackage{xcolor}
\usepackage[hang,flushmargin]{footmisc}
\usepackage{algorithm}
\usepackage{algpseudocode}
\usepackage{wrapfig}

\ifcamera \usepackage[accsupp]{axessibility} \fi

% ...

\ifarxiv  \fi

\newcommand{\R}[1]{{%
    \textbf{%
        \ifstrequal{#1}{1}{\textcolor{red}{R#1}}{%
        \ifstrequal{#1}{2}{\textcolor{blue}{R#1}}{%
        \ifstrequal{#1}{3}{\textcolor{magenta}{R#1}}{%
        \ifstrequal{#1}{4}{\textcolor{teal}{R#1}}{%
                           \textcolor{cyan}{R#1}%
        }}}}%
    }%
}}

\def\eqref#1{equation~\ref{#1}}

\def\1{\bm{1}}

\def\rd{{\mathrm{d}}}

\def\rvd{{\mathbf{d}}}

\def\rvw{{\mathbf{w}}}
\def\rvx{{\mathbf{x}}}
\def\rvy{{\mathbf{y}}}

\DeclareMathAlphabet{\mathsfit}{\encodingdefault}{\sfdefault}{m}{sl}
\SetMathAlphabet{\mathsfit}{bold}{\encodingdefault}{\sfdefault}{bx}{n}

\def\gN{{\mathcal{N}}}

\def\gT{{\mathcal{T}}}

\def\sR{{\mathbb{R}}}

\usepackage{xr-hyper}

\makeatletter
\newcommand*{\addFileDependency}[1]{
  \typeout{(#1)}
  \@addtofilelist{#1}
  \IfFileExists{#1}{}{\typeout{No file #1.}}
}

\makeatother

\definecolor{cvprblue}{rgb}{0.21,0.49,0.74}
\usepackage[pagebackref,breaklinks,colorlinks,allcolors=cvprblue]{hyperref}
\usepackage[capitalize]{cleveref}
\crefname{section}{Sec.}{Secs.}
\crefname{table}{Table}{Tables}
\crefname{figure}{Fig.}{Figs.}

\ifarxiv \crefname{appendix}{App.}{Apps.}
\else \crefname{appendix}{Suppl.}{Suppls.} \fi

\begin{document}
\newcommand{\ours}{\texttt{EPD-Solver}}
\newcommand{\oursplugin}{\texttt{EPD-Plugin}}

\newcommand{\tableCellHeight}{1}
\newcommand{\tabstyle}[1]{
  \setlength{\tabcolsep}{#1}
  \renewcommand{\arraystretch}{\tableCellHeight}
  \centering
  \small
}

\definecolor{tabhighlight}{HTML}{e5e5e5}
\definecolor{lightCyan}{rgb}{0.925,1,1}

\newtheorem{definition}{Definition}
\newtheorem{theorem}{Theorem}
\newtheorem{assumption}{Assumption}
\newtheorem{lemma}{Lemma}
\newtheorem{proposition}{Proposition}
\newtheorem{corollary}{Corollary}
\title{\paperTitle}
\author{\authorBlock}
\maketitle

\begin{abstract}
Diffusion models (DMs) have achieved state-of-the-art generative performance but suffer from high sampling latency due to their sequential denoising nature. Existing solver-based acceleration methods often face image quality degradation under a low-latency budget.
In this paper, we propose the Ensemble Parallel Direction solver (dubbed as \ours), a novel ODE solver that mitigates truncation errors by incorporating multiple parallel gradient evaluations in each ODE step. Importantly, since the additional gradient computations are independent, they can be fully parallelized, preserving low-latency sampling.
 Our method optimizes a small set of learnable parameters in a distillation fashion, ensuring minimal training overhead.
 In addition, our method can serve as a plugin to improve existing ODE samplers.
Extensive experiments on various image synthesis benchmarks demonstrate the effectiveness of our \ours~in achieving high-quality and low-latency sampling. For example, at the same latency level of 5 NFE, EPD achieves an FID of 4.47 on CIFAR-10, 7.97 on FFHQ, 8.17 on ImageNet, and 8.26 on LSUN Bedroom, surpassing existing learning-based solvers by a significant margin. Codes are
available in \url{https://github.com/BeierZhu/EPD}.
\end{abstract}
\section{Introduction}
\label{sec:intro}
\begin{figure}[t]
    \centering
\includegraphics[width=0.45\textwidth]{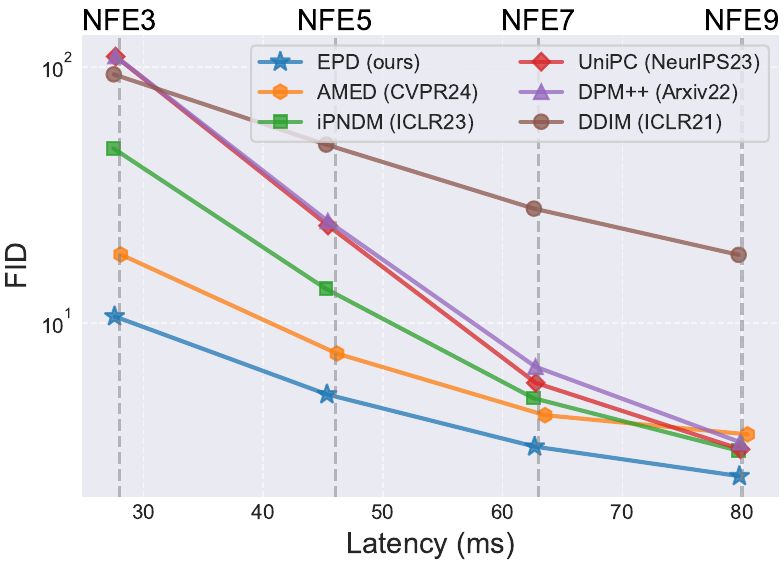}
    \caption{Comparison of various solvers on diffusion models. We compare the FID versus latency (ms) across different NFE settings on a NVIDIA 4090. Our proposed $\ours$ shows superior image quality without increasing latency.} 
    \label{fig:fidvslatency}
        \vspace{-4mm}
\end{figure}

Diffusion models (DMs)~\cite{sohl2015deep,ho2020denoising,rombach2022high} have become a leading paradigm in generative modeling, achieving state-of-the-art performance across a diverse range of applications, including image synthesis~\cite{rombach2022high,saharia2022photorealistic,Lei_2025_CVPR}, video generation~\cite{blattmann2023align,ho2022video}, speech synthesis~\cite{kong2021diffwave}, and 3D shape modeling~\cite{luo2021diffusion}. These models operate by gradually refining a noisy input through a denoising process, producing high-fidelity outputs with impressive diversity and realism. However, the multi-step sequential denoising process introduces substantial latency, making sampling inefficient.

In response to the challenge, recent efforts have focused on accelerating the sampling process of DMs. Notably, these methods typically fall into three categories: solver-based methods, distillation-based methods, and parallelism-based methods, each with distinct advantages and limitations. Solver-based methods develop fast numerical solvers to reduce sampling steps~\cite{songdenoising,karras2022elucidating,lu2022dpm,lu2022dpm_plus,liupseudo,zhangfast,zhao2024unipc,zhou2024fast,kim2024distilling,watson2021learning}. However, inherent truncation errors lead to significant quality degradation when the number of function evaluations (NFE) is low (\eg, $<5$). Distillation-based methods train a student DM to establish a bijective mapping between the data distribution and a predefined tractable noise distribution~\cite{zhou2025simple,luhman2021knowledge,liuflow,berthelot2023tract,salimansprogressive,meng2023distillation,song2023consistency,luo2023latent,kimconsistency}.
 This process allows the distilled model to generate high-quality samples within a minimal number of NFEs, often as low as one. However, achieving this level of efficiency requires extensive training with carefully designed objectives, making the distillation process computationally expensive. Additionally, such methods struggle to leverage multi-NFE settings effectively, limiting their flexibility when a trade-off between speed and quality is desired.
Parallelism-based methods accelerate diffusion models by trading computation for speed~\cite{shih2023parallel,li2024faster,li2024distrifusion,chenasyncdiff}. While promising, this direction remains underexplored.

\begin{figure*}[t]
    \centering
\includegraphics[width=0.95\textwidth]{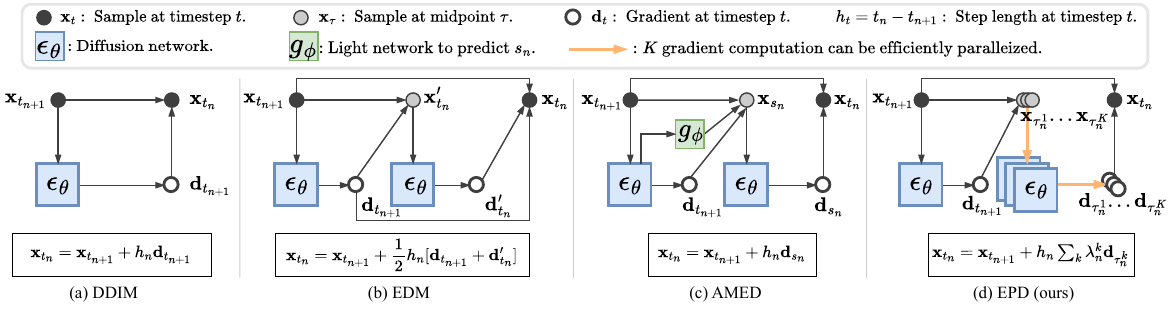}
    \caption{Computation graphs of various ODE solvers. \textbf{(a)} DDIM solver~\cite{songdenoising} (Euler's method) adopts the rectangle rule that uses the gradient at the start point: $\rvd_{t_{n+1}}=\bm{\epsilon}_\theta(\rvx_{t_{n+1}},t_{n+1})$. disclose EDM solver~\cite{karras2022elucidating} (Heun's method) uses the trapezoidal rule that averages the gradients of both the start and the end timesteps, \ie, $\rvd_{t_{n+1}}=\bm{\epsilon}_\theta(\rvx_{t_{n+1}},t_{n+1})$ and $\rvd_{t_{n}}'=\bm{\epsilon}_\theta(\rvx_{t_{n}}',t_{n})$, where $\rvx_{t_n}'$ is the additional evaluation given by Euler's method. \textbf{(c)} AMED solver~\cite{zhou2024fast} optimizes a small network $g_\phi(\cdot)$ to output an intermediate timestep $s_n \in (t_n,t_{n+1})$ to compute the gradient: $\rvd_{s_n}=\bm{\epsilon}_\theta(\rvx_{s_{n}},s_{n})$. Since AMED introduces a network in sequential computation, its latency is slightly higher than that of other solvers, as shown in~\cref{fig:fidvslatency}. \textbf{(d)} Our $\ours$ leverage $K$ parallel gradients to achieve more accurate integral approximation. We optimize $K$ intermediate timesteps $\tau_n^1, \dots, \tau_n^K$, compute their gradients $\rvd_{\tau_{n}^1}, \dots, \rvd_{\tau_{n}^K}$, and combine them via a simplex-weighted sum.} 
    \label{fig:teaser}
    \vspace{-2mm}
\end{figure*}

To combine the advantages of these approaches, we investigate solver-based methods under low-latency constraints and explore how additional computation can enhance image quality while maintaining minimal latency. 
We develop an Ensemble Parallel Direction (EPD) solver, which incorporates additional parallel gradient computations to mitigate truncation error in each ODE step.
At a high level, various existing ODE solvers utilize gradients at different timesteps to approximate the ODE solution with varying accuracy. For instance, as shown in~\cref{fig:teaser}, EDM~\cite{karras2022elucidating} (\cref{fig:teaser}.b) and AMED (\cref{fig:teaser}.c) improve image generation quality compared to DDIM (\cref{fig:teaser}.a) by leveraging additional gradients evaluated at $t_n$ and $s_n \in (t_{n+1}, t_n)$, respectively. Our EPD solver (\cref{fig:teaser}.d) extends this idea by incorporating $K$  learned intermediate timesteps ($\tau_n^k \in (t_{n+1}, t_n), k \in [K]$).
Combining these additional gradients via simplex-weighted summation yields a more accurate integral estimate, reducing local truncation error and enhancing sampling fidelity. 
Furthermore, since the computations of these additional gradients are independent -- each computed via a one-step Euler update from $\rvx_{t_{n+1}}$ -- they can be efficiently parallelized, ensuring no increase in inference latency. In~\cref{fig:fidvslatency}, we compare FID scores against latency for various ODE solvers on CIFAR-10~\cite{krizhevsky2009learning}. At each latency level, our $\ours$ with $K=2$ consistently achieves superior image quality. 

We optimize the learnable parameters (\eg, $\{\tau_n^k\}_{k=1}^K$) of our $\ours$ in a 
distillation fashion. Since the parameter count is small (ranging from 6 to 45 in our experiments), the tuning overhead remains minimal.
We further extend our method as a plugin to existing ODE samplers, termed \oursplugin.
We evaluate $\ours$ on a diverse set of image generation models, including CIFAR-10 \cite{krizhevsky2009learning}, FFHQ \cite{karras2019style}, ImageNet \cite{russakovsky2015imagenet}, LSUN Bedroom \cite{yu2015lsun}, and Stable Diffusion \cite{rombach2022high}. At the same latency level of 5 NFE, $\ours$ achieves an FID of 4.47 on CIFAR-10, 7.97 on FFHQ, 8.17 on ImageNet, and 8.26 on LSUN Bedroom.
This performance outperforms other learning-based solvers by a significant margin; for example, AMED Solver~\cite{zhou2024fast} only achieves an FID of 13.20 on LSUN Bedroom.
Our contributions can be summarized as follows:
\begin{itemize}
    \item We propose $\ours$, a novel ODE solver that leverages multiple parallel gradients to reduce truncation errors.
    \item We propose $\oursplugin$, a plugin that extends parallel gradient estimation to existing ODE samplers.
    \item With few learnable parameters, our solver is lightweight to train and does not increase inference latency.
    \item $\ours$ significantly outperforms existing ODE solvers in FID across multiple generation benchmarks.
\end{itemize}

\section{Related Work}
\label{sec:related}
High latency in the sampling process is a major drawback of DMs compared to other generative models~\cite{goodfellow2014generative,kingma2013auto}. Prior acceleration efforts mainly fall into three classes:

\noindent\textbf{Distillation-based methods.}
These methods accelerate diffusion models by re-training or fine-tuning the entire DM. One category is trajectory distillation, which trains a student model to imitate the teacher's trajectory with fewer steps~\cite{zhou2025simple}. This process can be achieved through offline distillation~\cite{luhman2021knowledge,liuflow}, which requires constructing a dataset sampled from teacher models, or online distillation, which progressively reduces sampling steps in a multi-stage manner~\cite{berthelot2023tract,salimansprogressive,meng2023distillation}. Another line of research is consistency distillation, where the denoising outputs along the sampling trajectory are enforced to remain consistent~\cite{song2023consistency,luo2023latent,kimconsistency}. Apart from distilling noise-image pairs, distribution matching methods match real and reconstructed samples at the distribution level~\cite{pooledreamfusion,wang2023prolificdreamer,sauer2024adversarial,yin2024one}. Despite significantly enhancing quality, these approaches incur high training costs and require carefully designed training procedures.

\noindent\textbf{Solver-based methods.}  
Beyond fine-tuning DMs, fast ODE solvers have been extensively studied. Training-free methods include Euler's method~\cite{songdenoising}, Heun’s method~\cite{karras2022elucidating}, Taylor expansion-based solvers (DPM-Solver~\cite{lu2022dpm}, DPM-Solver++~\cite{lu2022dpm_plus}), multi-step methods (PNDM~\cite{liupseudo}, iPNDM~\cite{zhangfast}), and predictor-corrector frameworks (UniPC~\cite{zhao2024unipc}). Some solvers require additional training, \eg, AMED-Solver~\cite{zhou2024fast}  
, D-ODE~\cite{kim2024distilling}, and DDSS~\cite{watson2021learning}. Recent work optimizes timestep schedules, with notable studies including LD3~\cite{tong2024learning}, AYS~\cite{sabour2024align}, GITS~\cite{chen2024trajectory}, and DMN~\cite{xue2024accelerating}.  
Though $\ours$ falls into this category, we optimize solver parameters via distillation to achieve high-quality, low-latency generation through parallelism. With minimal learnable parameters, training remains highly efficient.

\noindent\textbf{Parallelism-based methods.}
While promising, parallelism remains an underexplored approach for accelerating diffusion models. ParaDiGMS~\cite{shih2023parallel} leverages Picard iteration for parallel sampling but struggles to maintain consistency with original outputs. Faster Diffusion~\cite{li2024faster} performs decoder computation in parallel by omitting encoder computation at some adjacent timesteps, but this compromises image quality. Distrifusion~\cite{li2024distrifusion} divides high-resolution images into patches and performs parallel inference on each patch. AsyncDiff~\cite{chenasyncdiff} implements model parallelism through asynchronous denoising.  
Unlike prior methods that focus on reducing latency, our $\ours$ leverages parallel gradients to enhance image quality without incurring notable latency. 
\section{Method}
\label{sec:method}

\subsection{Background}
\label{sec:background}

Diffusion models gradually inject noise into data via a forward noising process and generate samples by learning a reversed denoising process, initialized with Gaussian noise. Let $\rvx \sim p_\text{data}(\rvx)$ denote the $d$-dimensional data and $p(\rvx;\sigma)$ the data distribution with Gaussian noise of variance $\sigma^2$ injected. The forward process is controlled by a noise schedule defined by the time scaling $s(t)$ and the noise level $\sigma(t)$ at time $t$.
In particular, $\rvx=s(t)\hat{\rvx}_t$, where $\hat{\rvx}_t \sim p(\rvx;\sigma(t))$. Such forward process can be formulated by a SDE~\citep{karras2022elucidating}:
\begin{equation}\label{eq:sde}
    \rd\rvx = \frac{\dot{s}(t)}{s(t)}\rvx + s(t)\sqrt{2\sigma(t)\dot{\sigma}(t)}\rd \rvw_t,
\end{equation}
where $\rvw \in \sR^d$ denotes Wiener process. In this paper, we adopt the framework of~\citet{karras2022elucidating} by setting $\sigma(t)=t$ and $s(t)=1$.
Generation is then performed with the reverse of~\cref{eq:sde}. Notably, there exists the probability flow ODE: 
\begin{equation}\label{eq:full-ode}
\rd\rvx =  - t\nabla_\rvx \log p(\rvx;t)\rd t 
\end{equation}
We learn a parameterized network $\bm{\epsilon}_\theta(\rvx,t)$ to predict the Gaussian noise added to $\rvx$ at time $t$. The network satisfies: $\bm{\epsilon}_\theta(\rvx,t)=-t\nabla_\rvx\log p(\rvx;t)$ and \cref{eq:full-ode} simplifies to: 
\begin{equation}\label{eq:ode-simple}
   \rd\rvx=\bm{\epsilon}_\theta(\rvx,t)\rd t
\end{equation}
The noise-prediction model $\bm{\epsilon}_\theta(\rvx,t)$ is trained by minimizing the $\ell_2^2$ loss with a weighting function $\lambda(t)$~\cite{karras2022elucidating,song2021scorebased}:
\begin{equation}
    \mathcal{L}_t(\theta)=\lambda(t)\mathbb{E}_{\rvx\sim p_\text{data},\bm{\epsilon}\sim \mathcal{N}(0,\mathbf{I})}\|\bm{\epsilon}_\theta(\rvx,t)-\bm{\epsilon}\|_2^2
\end{equation}
Given a time schedule $\gT=\{t_0=t_{\min},...,t_N=t_{\max}\}$, data generation involves starting from random noise $\rvx_{t_N} \sim \gN(\bm{0}, t_{\max}^2\mathbf{I})$, then iteratively solving \cref{eq:ode-simple} to compute the sequence $\{\rvx_{t_{N-1}},...,\rvx_{t_0}\}$. 

\subsection{The Proposed Solver}

\noindent\textbf{Motivation.} The solution of \cref{eq:ode-simple}
at time $t_n$ can be exactly computed in the integral form:
\begin{equation}
    \rvx_{t_n} = \rvx_{t_{n+1}} + \int_{t_{n+1}}^{t_n} \bm{\epsilon}_\theta(\rvx_t,t)\rd t
\end{equation}
Various ODE solvers have been proposed to approximate the integral. At a high level, these solvers leverage one or several points to compute gradients, which are then used to estimate the integral. Let $I$ denote the integral $I=\int_{t_{n+1}}^{t_n}\bm{\epsilon}_\theta(\rvx_t,t)\rd t$ and $h_n$ denote the step length $h_n=t_n-t_{n+1}$. For instance, DDIM~\cite{songdenoising} (Euler's method) adopts the rectangle rule that uses the gradient at the start point:
\begin{equation}
I \approx h_n \underbrace{\bm{\epsilon}_\theta(\rvx_{t_{n+1}},t_{n+1})}_{\text{start point grad.}}. 
\end{equation}
EDM~\cite{karras2022elucidating} considers the trapezoidal rule that averages the gradients of both the start and end points.
\begin{equation}
I \approx \frac{1}{2} h_n \{\underbrace{\bm{\epsilon}_\theta(\rvx_{t_{n+1}},t_{n+1})}_{\text{start point grad.}}+\underbrace{\bm{\epsilon}_\theta(\rvx'_{t_{n}},t_{n})}_{\text{end point grad.}}\}, 
\end{equation}
where $\rvx_{t_n}'$ is the additional evaluation point given by Euler's method, \ie,  $\rvx_{t_n}'=\rvx_{t_{n+1}}+h_n\bm{\epsilon}_\theta(\rvx_{t_{n+1}},t_{n+1})$. AMED-Solver~\cite{zhou2024fast} optimizes a small network to output an intermediate timestep $s_n \in (t_n,t_{n+1})$ to compute the gradient:
\begin{equation}
I \approx  h_n \underbrace{\bm{\epsilon}_\theta(\rvx_{s_n},s_n)}_{\text{midpoint grad.}}, 
\end{equation}
where $\rvx_{s_n}=\rvx_{t_{n+1}}+(s_n-t_{n+1})\bm{\epsilon}_\theta(\rvx_{t_{n+1}},t_{n+1})$. The computational graphs of DDIM, EDM, and AMED-Solver, illustrating their respective integral approximation processes, are shown in~\cref{fig:teaser}.

Compared to DDIM, EDM and AMED introduce an additional timestep for gradient computation ($t_n$ and $s_n$), leading to improved integral estimation. The key motivation behind our method is to leverage multiple timesteps to reduce the truncation errors. Furthermore, since the computations of additional gradients are independent, they can be efficiently parallelized without increasing inference latency. In this work, we propose the Ensemble Parallel Direction (EPD) solver, which refines the integral estimation by incorporating multiple intermediate timesteps. Formally, the integral is approximated as:
\begin{equation}\label{eq:epd1}
I \approx h_n \underbrace{\sum_{k=1}^K  \lambda^k_n\bm{\epsilon}_\theta(\rvx_{\tau^k_n},\tau^k_n)}_{\text{ensemble parallel grads.}},
\end{equation}
where $\tau^k_n \in (t_n,t_{n+1})$ are the intermediate timesteps, and the weights form a simplex combination satisfying $\lambda^k_n \geq 0$ and $\sum_{k=1}^K \lambda^k_n = 1$. The state at each intermediate timestep $\tau^k_n$ is computed using Euler’s method as: $\rvx_{\tau^k_n}=\rvx_{t_{n+1}}+(\tau_k-t_{n+1})\bm{\epsilon}_\theta(\rvx_{t_{n+1}},t_{n+1})$. Each gradient computation $\bm{\epsilon}_\theta(\rvx_{\tau^k_n},\tau^k_n)$ is fully parallelizable, preserving efficiency without increasing inference latency. 
In fact, the use of gradients estimated at multiple timesteps for improved integral approximation can be theoretically justified by the following mean value theorem for vector-valued functions.
\begin{theorem}\label{theorem:1}
(\cite{mcleod1965mean})
    When $f$ has values in an $n$-dimensional vector space and is continuous on the closed interval $[a,b]$ and differentiable on the open interval $(a,b)$, we have
    \begin{equation}
        f(b)-f(a)=(b-a)\sum_{k=1}^n \lambda_k f'(c_k),
    \end{equation}
    for some $c_k\in (a,b), \lambda_k \geq 0$, and $\sum_{k=1}^n\lambda_k=1$.
\end{theorem}
In the context of denoising process, the  function outputs an $d$-dimensional vector as $\rvx \in \mathbb{R}^d$. 
According to~\cref{theorem:1}, the exact integral of $\bm{\epsilon}_\theta(\rvx_t,t)$ over the interval $[t_n,t_{n+1}]$ can be expressed as a simplex-weighted combination of  gradients evaluated at $d$ intermediate points, scaled by the interval length $h_n=t_n-t_{n+1}$, as formulated in~\cref{eq:epd1}.

\noindent\textbf{Parameters optimizing and inference.}
\cite{ningelucidating,li2024alleviating} identify exposure bias—\ie, the mismatch between training and sampling inputs—as a key factor contributing to error accumulation and sampling drift. To mitigate this, they propose scaling the network output and shifting the timestep, respectively. Inspired by these insights, we introduce two learnable parameters, $o_n$ and $\delta_n^k$, to perturb the scale of network output’s and the timestep. Our $\ours$ follows the update rule: 
\begin{equation}\label{eq:epdsolverfinal}
\rvx_{t_n} = \rvx_{t_{n+1}} + (1+o_n)h_n\sum_{k=1}^K  \lambda^k_n\bm{\epsilon}_\theta(\rvx_{\tau^k_n},\tau^k_n+\delta^k_n)
\end{equation}
We define the parameters at step $n$ as ${\Theta}_n = \{\tau^k_n, \lambda_n^k, \delta^k_n, o_n \}_{k=1}^K$ and denote the complete set of parameters for an $N$-step sampling process as ${\Theta}_{1:N}$. Consequently, the total number of parameters is given by $N(1+3K)$.

To determine ${\Theta}_{1:N}$, we employ a distillation-based optimization process. 
Specifically, given a student time schedule with $N$ steps $\mathcal{T}_{\mathsf{stu}}=\{t_0=t_{\min},...,t_N=t_{\max}\}$, we insert $M$ intermediate steps between $t_n$ and $t_{n+1}$, \ie, $\mathcal{T}_{\mathsf{tea}}=\{t_0,...,t_n,t_n^1,...,t_n^M,t_{n+1},..,t_N\}$, to yield a more accurate teacher trajectories.
The training process starts with generating teacher trajectories by any ODE solver (\eg, DPM-Solver) and store the reference states as $\{\rvy_{t_n}\}_{n=0}^N$. Afterward, we sample student trajectory with the same initial noise $\rvy_{t_N}$, and optimize the parameters $\{\Theta_n\}_{n=1}^N$ to obtain the student trajectory $\{\rvx_{t_n}\}_{n=0}^N$ that aligns the teacher trajectory w.r.t some distance measurement $\text{dist}(\cdot,\cdot)$. For noisy states $\{\rvx_{t_n}\}_{n=1}^N$, we use the squared $\ell_2$ distance as $\text{dist}(\cdot,\cdot)$. For a generated sample $\rvx_{t_0}$, we compute the squared $\ell_2$ distance in the feature space of the last layer of an ImageNet-pretrained Inception network~\cite{szegedy2015going}.
In particular, to improve the alignment between $\rvx_{t_n}$ and $\rvy_{t_n}$, since the value of $\rvx_{t_n}$ is dependent of the parameters $\Theta_1$ to $\Theta_n$, we aim to optimize them by minimizing
\begin{equation}\label{eq:loss}
    \mathcal{L}_n(\Theta_{1:n})=\text{dist}(\rvx_{t_n}, \rvy_{t_n}).
\end{equation}
In one training loop, we require $N$ backpropagation. The entire training algorithm is listed in~\cref{algo:train} and the inference procedure is provided in~\cref{algo:sample}.
By default, we adopt the analytical first step (AFS) trick~\cite{dockhorn2022genie} in the first step to save one NFE by simply using $\rvx_{t_N}$ as direction. 

\begin{algorithm}[t]
\caption{Optimizing $\Theta_{1:N}$}\label{algo:train}
\begin{algorithmic}[1]
\State \textbf{Given:} Time schedules $\mathcal{T}_\mathsf{stu}$ and $\mathcal{T}_\mathsf{tea}$, teacher solver $\mathcal{S}$.
\State \textbf{Return:} $\Theta_{1:N}$, where $\Theta_n=\{\tau_n^k,\lambda_n^k,\delta_n^k,o_n\}_{k=1}^K$
\Repeat 
\State Initialize $\rvx_{t_N}=\rvy_{t_N}\sim \mathcal{N}(\mathbf{0},t^2_N\mathbf{I})$
\State Sample a teacher trajectory $\{\rvy_{t_n}\}_{n=1}^N$ via $\mathcal{S}$
\For{$n = N-1$ \textbf{to} $0$}
    \State Compute $\rvx_{t_n}$ using~\cref{eq:epdsolverfinal}
    \State Update $\Theta_{1:n}$ via $\min\mathcal{L}_n(\Theta_{1:n})$ (~\cref{eq:loss})
\EndFor

\Until{converge}
\end{algorithmic}
\end{algorithm}
\begin{algorithm}[t]
\caption{$\ours$ sampling}\label{algo:sample}
\begin{algorithmic}[1]
\State \textbf{Given:} Time schedule $\mathcal{T}_\mathsf{stu}$, learned parameters $\Theta_{1:N}$.
\State \textbf{Return:} $\rvx_{t_0}$
\State Initialize $\rvx_{t_N}\sim \mathcal{N}(\mathbf{0},t^2_N\mathbf{I})$
\For{$n = N-1$ \textbf{to} $0$}
    \State $I\leftarrow  (1+o_n)h_n\sum_{k=1}^K  \lambda_n^k\bm{\epsilon}_\theta(\rvx_{\tau_n^k},\tau^k_n+\delta^k_n)$
    \Statex {\color{blue} \Comment{implement parallelism for accelerating}}
    \State $\rvx_{t_n} \leftarrow  \rvx_{t_{n+1}} + I$
\EndFor
\end{algorithmic}
\end{algorithm}
\noindent\textbf{\oursplugin~to existing solvers.} 
\ours~can be applied to existing solvers to further enhance diffusion sampling. The key idea is to replace their original gradient estimation with multiple parallel branches. As a representative case, we demonstrate this using the multi-step iPNDM sampler~\cite{liupseudo,zhangfast}.
We refer to the modified solver as \oursplugin. Due to space limitations, a detailed description is deferred to \cref{sec:intro-plugin}.

\subsection{Discussion}
\noindent\textbf{Discussion with multi-step solvers.} While multi-step solvers~\cite{lu2022dpm_plus,zhao2024unipc,liupseudo,zhangfast} also use multiple gradients to approximate the integral, they typically rely on Taylor expansion or polynomial extrapolation to linearly combine \textit{historical} gradients. In contrast, our method is grounded in the vector-valued mean value theorem and optimizes a convex combination of gradients evaluated \textit{within the current time interval}. By focusing on in-interval gradients, our approach yields a more accurate and adaptive approximation of the integral.

\begin{figure}[t]
    \centering
    \includegraphics[width=0.45\textwidth]{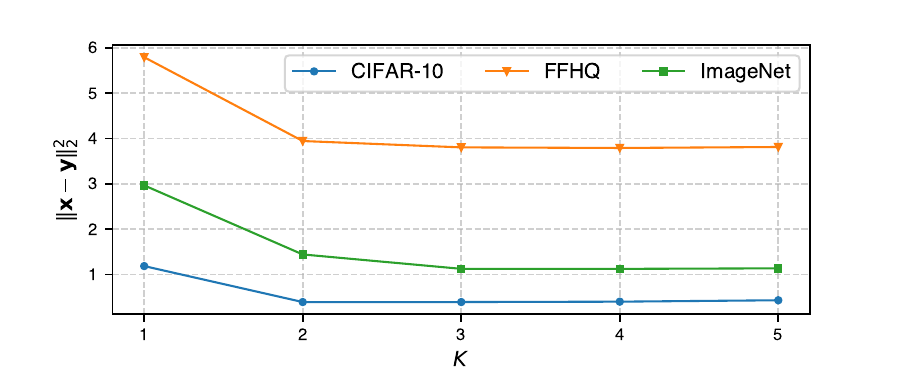}
    \caption{$\ell_2$ error between teacher and student trajectory \wrt $K$.} 
    \label{fig:trun_error}
    \vspace{-4mm}
\end{figure}

\noindent\textbf{Discussion with AMED-Solver.}
\noindent AMED-Solver~\cite{zhou2024fast} estimates the direction using a single intermediate timestep per step. In contrast, our $\ours$ method combines multiple intermediate gradients via a convex weighting scheme, without increasing inference latency. While a single direction may suffice when the trajectory is nearly one-dimensional, PCA analysis in~\cite{chen2024trajectory} shows that the first principal component accounts for only 65\% of the variance, suggesting that multiple directions better capture the underlying geometry.

To verify this, we conduct a controlled experiment with a 3-step schedule. As shown in~\cref{fig:trun_error}, we compute the $\ell_2$ error between teacher and student trajectories over 1000 random samples, varying the number of intermediate gradients $K$. The error drops significantly from $K=1$ to $K=2$, but shows diminishing returns for $K>2$, indicating that two directions already capture most of the trajectory's structure.

In addition, unlike AMED-Solver, which uses a neural network to predict sample-specific interpolation points, our $\ours$ learns global sampling parameters in a plug-and-play fashion, without incurring extra runtime cost.
\section{Experiments}
\label{sec:exp}
This section is organized as follows:
\begin{itemize}
    \item \cref{sec:exp_setup} provides an overview of our experimental setup.
    \item \cref{sec:main_results} compares our $\ours$ and $\oursplugin$ with state-of-the-art ODE samplers.
    \item \cref{sec:numKanalysis} analyzes the impact of the number of parallel directions $K$ on image quality and inference latency.
    \item \cref{sec:ablations} ablates the main components of $\ours$.
    \item \cref{sec:qualitative} showcases qualitative visualizations of the sampling process and generated images.
\end{itemize}

\subsection{Setup}\label{sec:exp_setup}

\noindent\textbf{Models.}
We test out ODE solvers on diffusion-based image generation models, covering both pixel-space~\cite{karras2022elucidating} and latent-space models~\cite{rombach2022high}, across image resolutions ranging from 32 to 512.
For pixel-space models, we evaluate the pretrained models on CIFAR 32$\times$32~\cite{krizhevsky2009learning}, FFHQ 64$\times$64~\cite{karras2019style}, ImageNet 64$\times$64~\cite{russakovsky2015imagenet} from~\cite{karras2022elucidating}. 
For latent-space models, we examine 
the pretrained models on LSUN Bedroom 256$\times$256~\cite{yu2015lsun} from~\cite{rombach2022high} and Stable-Diffusion~\cite{rombach2022high}  at a resolution of 512.

\noindent\textbf{Baseline solvers.}
We compare against representative ODE solvers across three categories:
(1) Single-step solvers: DDIM~\cite{songdenoising}, EDM~\cite{karras2022elucidating}, DPM-Solver-2~\cite{lu2022dpm}, and AMED-Solver~\cite{zhou2024fast};
(2) Multi-step solvers: DPM-Solver++(3M)\cite{lu2022dpm_plus}, UniPC\cite{zhao2024unipc}, iPNDM~\cite{liupseudo,zhangfast}, and AMED-Plugin~\cite{zhou2024fast};
(3) Parallelism-based solver: ParaDiGMS~\cite{shih2023parallel}. 
For a fair comparison, we follow the recommended time schedules from their original papers~\cite{karras2022elucidating,lu2022dpm_plus,zhao2024unipc}. Specifically, we use the logSNR schedule for DPM-Solver-2, DPM-Solver++(3M), and UniPC, the time-uniform schedule for AMED-Solver~\cite{zhou2024fast}, while employing the polynomial time schedule with $\rho=7$ for the remaining baselines. Please refer to~\cref{sec:paradigm_details} for implementation details of ParaDiGMS~\cite{shih2023parallel}.

\noindent\textbf{Evaluation.} 
We test our \texttt{EPD-\{Solver, Plugin\}}~under low NFE budgets ($\text{NFE}\in \{3,5,7,9\}$) where AFS~\cite{dockhorn2022genie} is applied. \texttt{EPD-\{Solver, Plugin\}} have the same NFE as the baselines when $K=1$. For $K>1$, each step involves $K-1$ extra NFE. However, parallelism ensures that latency remains unchanged. We use the term Parallel NFE (Para. NFE) to denote the effective NFE under parallel execution.
We assess sample quality using the Fréchet Inception Distance (FID) computed over 50k images.
For Stable-Diffusion, we evaluate FID by generating 30k images using prompts sampled from the MS-COCO validation set~\cite{lin2014microsoft}.

\noindent\textbf{Implementation details.}
We optimize our parameters using the Adam optimizer on 10k images with a batch size of 32.
To prevent overfitting, we constrain $o_n$ and $\delta_n^k$ using the sigmoid trick, ensuring they remain within $[-0.05, 0.05]$. 
Since the parameter count is small (ranging from 6 to 45 in our experiments), training is highly efficient—taking $\sim$3 minutes for CIFAR-10 on a single NVIDIA 4090 and $\sim$30 minutes for LSUN Bedroom 256$\times$256 on four NVIDIA A800 GPUs. To generate teacher trajectory, we employ DPM-Solver-2 solver with $M=6$ intermediate time steps injected.
Additional implementation details are available in \cref{sec:implement-epd}.
\subsection{Main Results}\label{sec:main_results}

\makeatletter
\renewcommand\thesubtable{(\alph{subtable})}
\makeatother

\begin{table*}[t!]
\small 
\captionsetup[subfloat]{labelformat=simple, labelsep=space}
\begin{minipage}[t]{0.48\textwidth}
\fontsize{8}{10}\selectfont
\subfloat[Unconditional generation on \textbf{CIFAR10} $32 \times 32$ \cite{krizhevsky2009learning}]{
\begin{tabular}{llcccc}
\toprule
 &\multirow{2}{*}{Method} & \multicolumn{4}{c}{(Para.) NFE} \\
\cmidrule{3-6} & & 3 & 5 & 7 & 9 \\
\midrule
\multirow{4}{*}{\rotatebox{90}{Single-step}} & DDIM~\cite{songdenoising} & 93.36 & 49.66 & 27.93 & 18.43 \\
&EDM~\cite{karras2022elucidating} & 306.2 & 97.67 & 37.28 & 15.76 \\
&DPM-Solver-2~\cite{lu2022dpm} & 155.7 & 57.30 & 10.20 & 4.98 \\
&AMED-Solver~\cite{zhou2024fast} & 18.49 & 7.59 & 4.36 & 3.67 \\ \midrule
\multirow{4}{*}{\rotatebox{90}{Multi-step}} 
& DPM-Solver++(3M)~\cite{lu2022dpm_plus} & 110.0 & 24.97 & 6.74 & 3.42 \\
&UniPC~\cite{zhao2024unipc} & 109.6 & 23.98 & 5.83 & 3.21 \\
&iPNDM~\cite{liupseudo,zhangfast} & 47.98 & 13.59 & 5.08 & 3.17 \\ 
&AMED-Plugin~\cite{zhou2024fast} & 10.81 & 6.61 & 3.65 & 2.63 \\ 
\midrule
\multirow{3}{*}{\rotatebox{90}{Parallel}} &ParaDiGMS~\cite{shih2023parallel} & 51.03 & 18.96 & 7.18 & 6.19 \\
&\cellcolor{lightCyan}\ours~(ours) & \cellcolor{lightCyan}\textbf{10.40}  & \cellcolor{lightCyan}\textbf{4.33}  & \cellcolor{lightCyan}\textbf{2.82} & \cellcolor{lightCyan}\underline{2.49}\\
& \cellcolor{lightCyan}\oursplugin~(ours) & \cellcolor{lightCyan}\underline{10.54} & \cellcolor{lightCyan}\underline{4.47} & \cellcolor{lightCyan}\underline{3.27} & \cellcolor{lightCyan}\textbf{2.42} \\
\bottomrule
\end{tabular}
}
\vspace{0.5em} 
\subfloat[Unconditional generation on \textbf{FFHQ} $64 \times 64$ \cite{karras2019style}]{
\begin{tabular}{llcccc}
\toprule
 &\multirow{2}{*}{Method} & \multicolumn{4}{c}{(Para.) NFE} \\
\cmidrule{3-6} & & 3 & 5 & 7 & 9 \\
\midrule
\multirow{4}{*}{\rotatebox{90}{Single-step}} 
&DDIM~\cite{songdenoising} & 78.21 & 43.93 & 28.86 & 21.01 \\
&EDM~\cite{karras2022elucidating} & 356.5 &	116.7 	&54.51 &	28.86  \\
&DPM-Solver-2~\cite{lu2022dpm} & 266.0 & 87.10 & 22.59 & 9.26 \\
&AMED-Solver~\cite{zhou2024fast}  & 47.31 & 14.80 & 8.82 & 6.31 \\ \midrule
\multirow{4}{*}{\rotatebox{90}{Multi-step}} 
&DPM-Solver++(3M)~\cite{lu2022dpm_plus} & 86.45 & 22.51 & 8.44 & 4.77 \\
&UniPC~\cite{zhao2024unipc} & 86.43 & 21.40 & 7.44 & 4.47 \\ 
&iPNDM~\cite{liupseudo,zhangfast} & 45.98 & 17.17 & 7.79 & 4.58 \\ 
&AMED-Plugin~\cite{zhou2024fast}  & 26.87 & 12.49 & 6.64  & 4.24 \\\midrule
\multirow{3}{*}{\rotatebox{90}{Parallel}} 
&ParaDiGMS~\cite{shih2023parallel} & 43.64 & 20.92 & 16.39 & 8.81 \\
&\cellcolor{lightCyan}\ours~(ours) & \cellcolor{lightCyan}\underline{21.74}  & \cellcolor{lightCyan}\textbf{7.84} & \cellcolor{lightCyan}\textbf{4.81} & \cellcolor{lightCyan}\underline{3.82}\\
&\cellcolor{lightCyan}\oursplugin~(ours) &\cellcolor{lightCyan}\textbf{19.02}&\cellcolor{lightCyan}\underline{7.97}&\cellcolor{lightCyan}\underline{5.09}& \cellcolor{lightCyan}\textbf{3.53} \\
\bottomrule
\end{tabular}
}
\end{minipage}
\hfill
\begin{minipage}[t]{0.48\textwidth}
\centering
  \fontsize{8}{10}\selectfont

\subfloat[Conditional generation on \textbf{ImageNet} $64 \times 64$ \cite{russakovsky2015imagenet}]{
\begin{tabular}{llcccc}
\toprule
 &\multirow{2}{*}{Method} & \multicolumn{4}{c}{(Para.) NFE} \\
\cmidrule{3-6} & & 3 & 5 & 7 & 9 \\
\midrule
\multirow{4}{*}{\rotatebox{90}{Single-step}} 
&DDIM~\cite{songdenoising} & 82.96 & 43.81 & 27.46 & 19.27 \\
&EDM~\cite{karras2022elucidating} & 249.4 & 89.63 & 37.65 & 16.76 \\
&DPM-Solver-2~\cite{lu2022dpm} & 140.2 & 42.41 & 12.03 & 6.64 \\ 
&AMED-Solver~\cite{zhou2024fast} & 38.10 & 10.74 & 6.66 & 5.44 \\ \midrule
\multirow{4}{*}{\rotatebox{90}{Multi-step}} 
&DPM-Solver++(3M)~\cite{lu2022dpm_plus} & 91.52 & 25.49 & 10.14 & 6.48 \\
&UniPC~\cite{zhao2024unipc} & 91.38 & 24.36 & 9.57 & 6.34 \\
&iPNDM~\cite{liupseudo,zhangfast} & 58.53 & 18.99 & 9.17 & 5.91 \\
&AMED-Plugin~\cite{zhou2024fast} & 28.06  & 13.83  & 7.81  & 5.60 \\ \midrule
\multirow{3}{*}{\rotatebox{90}{Parallel}} 
&ParaDiGMS~\cite{shih2023parallel} & 41.11 & 17.27 & 13.67 & 6.38 \\
&\cellcolor{lightCyan}\ours~(ours) & \cellcolor{lightCyan}\textbf{18.28} & \cellcolor{lightCyan}\textbf{6.35} & \cellcolor{lightCyan}\underline{5.26} & \cellcolor{lightCyan}\underline{4.27}\\
&\cellcolor{lightCyan}\oursplugin~(ours) & \cellcolor{lightCyan}\underline{19.89} & \cellcolor{lightCyan}\underline{8.17} & \cellcolor{lightCyan}\textbf{4.81} & \cellcolor{lightCyan}\textbf{4.02} \\
\bottomrule
\end{tabular}
}

\vspace{0.5em}

\subfloat[Unconditional generation on \textbf{LSUN Bedroom} $256 \times 256$ \cite{yu2015lsun}]{
\begin{tabular}{llcccc}
\toprule
 &\multirow{2}{*}{Method} & \multicolumn{4}{c}{(Para.) NFE} \\
\cmidrule{3-6} & & 3 & 5 & 7 & 9 \\
\midrule
\multirow{4}{*}{\rotatebox{90}{Single-step}} 
&DDIM~\cite{songdenoising} & 86.13 & 34.34 & 19.50 & 13.26 \\
&EDM~\cite{karras2022elucidating} &
291.5	 & 175.7	 & 78.67	  & 35.67  \\
&DPM-Solver-2~\cite{lu2022dpm} & 210.6 & 80.60 & 23.25 & 9.61 \\
&AMED-Solver~\cite{zhou2024fast} & 58.21 & 13.20 & 7.10 & 5.65 \\ \midrule
\multirow{4}{*}{\rotatebox{90}{Multi-step}} 
&DPM-Solver++(3M)~\cite{lu2022dpm_plus} & 111.9 & 23.15 & 8.87 & 6.45 \\
&UniPC~\cite{zhao2024unipc} & 112.3 & 23.34 & 8.73 & 6.61 \\
&iPNDM~\cite{liupseudo,zhangfast} & 80.99 & 26.65 & 13.80 & 8.38 \\
&AMED-Plugin~\cite{zhou2024fast} & 101.5 & 25.68 & 8.63 & 7.82 \\ \midrule
\multirow{3}{*}{\rotatebox{90}{Parallel}} 
&ParaDiGMS~\cite{shih2023parallel} & 100.3 & 31.68 & 15.85 & 8.56 \\
&\cellcolor{lightCyan}\ours~(ours) & \cellcolor{lightCyan}\textbf{13.21} & \cellcolor{lightCyan}\textbf{7.52} & \cellcolor{lightCyan}\underline{5.97} & \cellcolor{lightCyan}\underline{5.01} \\
&\cellcolor{lightCyan}\oursplugin~(ours) & \cellcolor{lightCyan}\underline{14.12} & \cellcolor{lightCyan}\underline{8.26} & \cellcolor{lightCyan}\textbf{5.24} & \cellcolor{lightCyan}\textbf{4.51} \\
\bottomrule
\end{tabular}
}
\end{minipage}

\caption{
Image generation results across four datasets: 
{(a)} CIFAR10, 
{(b)} FFHQ, 
{(c)} ImageNet, 
{(d)} LSUN Bedroom. 
We compared our \ours~and \oursplugin~with (1) Single-step solvers: DDIM, EDM, DPM-Solver-2 and AMED-Solver, (2) Multi-step solvers: DPM-Solver++(3M), UniPC, iPNDM and AMED-Plugin, (3) Parallelism-based solver: ParaDiGMS. The best results are in \textbf{bold}, the second best are \underline{underlined}. See~\cref{sec:learnedparameters} for the value of the learned parameters of \ours~and~\oursplugin. 
}
\label{tab:main_results}
\vspace{-4mm}
\end{table*}

\begin{table}[t]
  \centering
      \fontsize{8}{10}\selectfont
      \setlength{\tabcolsep}{8pt} 
    \begin{tabular}{lcccc}
      \toprule
      \multirow{2}{*}{Method} & \multicolumn{4}{c}{(Para.) NFE} \\
      \cmidrule{2-5}
      & 8 & 12 & 16 & 20 \\
      \midrule
      DPM-Solver++(2M)~\cite{lu2022dpm_plus}       & 21.33 & 15.99 & 14.84 & 14.58 \\
      AMED-Plugin~\cite{zhou2024fast}             & {18.92} & {14.84} & {13.96} & {13.24} \\
      \rowcolor{lightCyan}
$\ours$ (ours) & \textbf{16.46}  & \textbf{13.14} & \textbf{12.52} & \textbf{12.17}\\
      \bottomrule
    \end{tabular}
  \caption{FID results on Stable-Diffusion~\cite{rombach2022high}.}
  \label{tab:sup_fid_stable_diff}
  \vspace{-4mm}
\end{table}

In~\cref{tab:main_results}, we compare the FID scores of images generated by our \ours~with $K=2$ against baseline solvers across the CIFAR-10, FFHQ, ImageNet, and LSUN Bedroom datasets. The results demonstrate consistent and significant improvements from our learned directions across all datasets and NFE values. Specifically, with 9 (Para.) NFE, we achieve FID scores of 4.27 and 5.01 on the ImageNet and LSUN datasets, respectively, while the second-best baseline counterpart achieves 5.44 and 5.65, showing a notable improvement. Moreover, in the low NFE region, such as 3 NFE on LSUN Bedroom, our $\ours$ achieves a remarkable 13.21 FID, significantly outperforming the second-best baseline solver (AMED-Solver), which achieves 58.21 FID. 
We further evaluate \oursplugin~applied to the iPNDM solver, and observe that it outperforms \ours~when $\text{NFE} > 7$, consistent with our expectation that iPNDM benefits from historical gradients only when the step is sufficiently large. With small NFE, this advantage is less pronounced.

We evaluate our $\ours$ method on Stable-Diffusion v1.5, setting the classifier-free guidance weight to 7.5, and report the FID score on the MS-COCO validation set in~\cref{tab:sup_fid_stable_diff}. Additionally, we compare the quality of samples generated by DPM-Solver(2M)++ (as recommended in the official implementation) and the AMED-Plugin Solver, a recent SoTA solver. The results demonstrate the consistent superiority of our proposed method.

\begin{figure*}[t]
    \centering
\includegraphics[width=1.0\textwidth]{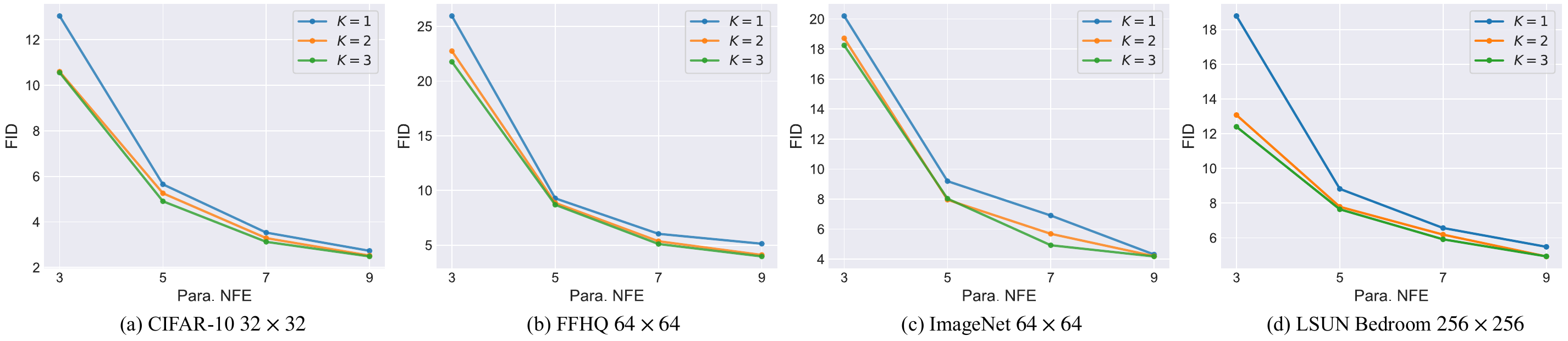}
    \caption{FID curves for different datasets and the number of parallel directions ($K$).} 
    \label{fig:fids_curve}
\end{figure*}

\begin{table*}[t!]
\small 
\captionsetup[subfloat]{labelformat=simple, labelsep=space}
\begin{minipage}[t]{0.48\textwidth}
    \fontsize{8}{10}\selectfont
    \subfloat[\textbf{CIFAR10} and \textbf{FFHQ}]{
        \begin{tabular}{lccccc}
\toprule
&  \multirow{2}{*}{$K$}  & \multicolumn{4}{c}{Para. NFE}  \\ \cmidrule{3-6}
                          &     & 3    & 5    & 7   & 9   \\ \midrule
\multirow{3}{*}{\rotatebox{90}{CIFAR}} 
& $1$ &  28.1$\pm$0.84 & 47.2$\pm$0.88 & 63.5$\pm$0.71 & 80.5$\pm$0.73                  \\
& $2$ &  27.6$\pm$0.78 & 45.3$\pm$0.77 & 62.7$\pm$0.76& 79.8$\pm$0.81  \\
 & $3$ &  27.7$\pm$0.85 & 45.7$\pm$0.80 &  63.5$\pm$0.86 & 82.0$\pm$0.94  \\ \midrule   
\multirow{3}{*}{\rotatebox{90}{FFHQ}}
& $1$ & 34.4$\pm$0.79 &  56.1$\pm$0.78  &  77.4$\pm$0.96 & 100.4$\pm$0.74                \\
 & $2$ & 34.4$\pm$0.85  &  56.4$\pm$0.83 &  79.6$\pm$0.92 & 98.6$\pm$0.83    \\
& $3$ & 34.1$\pm$0.92  &   56.0$\pm$0.88 & 78.0$\pm$0.89 & 99.8$\pm$0.94    \\ \bottomrule
\end{tabular}
        }
\end{minipage}\hfill
\begin{minipage}[t]{0.48\textwidth}
    \fontsize{8}{10}\selectfont
        \centering
    \subfloat[ \textbf{ImageNet} and \textbf{LSUN Bedroom}]{
        \begin{tabular}{lccccc}
\toprule
&  \multirow{2}{*}{$K$}  & \multicolumn{4}{c}{Para. NFE}  \\ \cmidrule{3-6}
                          &     & 3    & 5    & 7   & 9   \\ \midrule
\multirow{3}{*}{\rotatebox{90}{IN}} & $1$ & 56.7$\pm$1.09   & 93.3$\pm$1.04  & 128.2$\pm$1.06   &  163.2$\pm$1.08                  \\
& $2$ &  55.7$\pm$1.16  &   92.3$\pm$1.18 & 128.2$\pm$1.14 & 164.4$\pm$1.23   \\
& $3$ & 55.7$\pm$1.20  &   94.7$\pm$1.20     &129.9$\pm$1.21 & 162.8$\pm$1.20    \\ \midrule   
\multirow{3}{*}{\rotatebox{90}{LSUN}} 
& $1$ & 57.5$\pm$1.26  & 78.8$\pm$1.02 & 104.3$\pm$1.15  & 131.1$\pm$1.03                 \\
& $2$ &  56.6$\pm$1.16 & 82.6$\pm$1.12 & 109.6$\pm$1.10 &  138.9$\pm$1.23 \\
& $3$ &  57.9$\pm$1.15 & 86.2$\pm$1.16 & 117.8$\pm$1.10 & 147.8$\pm$1.19  \\ \bottomrule 
\end{tabular}
        }
\end{minipage}
\caption{Latency (ms) measured across different datasets, Para. NFE values, and the number of parallel directions ($K$). No noticeable latency increase was observed when $K$ increased to 2. The reported values include the 95\% confidence interval. }
\label{tab:latency}
\end{table*}

\begin{table*}[t]
    \centering
    \begin{minipage}[t]{0.33\textwidth}
        \centering
        \fontsize{8}{10}\selectfont
         \setlength{\tabcolsep}{5pt} 
        \begin{tabular}{lcccc}
        \toprule
         Para. NFE & 3 & 5 & 7 & 9 \\
        \midrule
        $\ours$ & \textbf{10.40} & \textbf{4.33} & \textbf{2.82} & \textbf{2.49} \\ 
        \quad w.o. $o_n$ &13.25 & 	5.84 &	3.59	& 2.79  \\
        \quad w.o. $\delta_n^k$ & 13.02	&5.47	 & 3.23	& 2.69 \\
        \quad w.o. $o_n$ \& $\delta_n^k$ &16.01	&6.62&	4.24&	3.24 \\
        \bottomrule
        \end{tabular}
        \caption{Effect of scaling factors ($o_n, \delta_n^k$).}\label{tab:scaling}   
    \end{minipage}
    \hfill
    \begin{minipage}[t]{0.33\textwidth}
        \centering
        \fontsize{8}{10}\selectfont
         \setlength{\tabcolsep}{5pt} 
        \begin{tabular}{lcccc}
        \toprule
        \multirow{2}{*}{Schedule} & \multicolumn{4}{c}{Para. NFE} \\
        & 3 & 5 & 7 & 9 \\
        \midrule
        LogSNR &54.07&8.88&7.95& 3.97\\
        EDM~\cite{karras2022elucidating} & 11.10 &8.89  & 4.50& 3.72 \\
        Time-uniform  & \textbf{10.40} & \textbf{4.33} & \textbf{2.82} & \textbf{2.49} \\
        \bottomrule
        \end{tabular}
        \caption{Effect of time schedules.}\label{tab:schedule}
    \end{minipage}
    \hfill
    \begin{minipage}[t]{0.33\textwidth}
         \centering
        \fontsize{8}{10}\selectfont 
         \setlength{\tabcolsep}{5pt} 
        \begin{tabular}{lcccc}
        \toprule
        \multirow{2}{*}{Teacher Solver} & \multicolumn{4}{c}{Para. NFE} \\
        & 3 & 5 & 7 & 9 \\
        \midrule
                Heun \cite{karras2022elucidating} & 15.91 & 6.65 & 4.61 & 3.57 \\
               iPNDM \cite{liupseudo,krizhevsky2009learning} & 13.69 & 6.64 & 4.59 & 3.59 \\
                DPM-Solver-2 \cite{lu2022dpm} & \textbf{10.40} & \textbf{4.33} & \textbf{2.82} & \textbf{2.49} \\
        \bottomrule
        \end{tabular}
        \caption{Effect of teacher ODE solvers.}
        \label{tab:teacher}      
    \end{minipage}
\end{table*}

\subsection{On the Number of Parallel Directions}\label{sec:numKanalysis}

\noindent\textbf{Image quality with different values of $K$.}
In~\cref{fig:fids_curve}, 
we compare the quality of images generated using our $\ours$ with different values of $K$. As expected, increasing the number of intermediate points leads to improved FID scores. For example, on the FFHQ dataset with 3 Para. NFE, the FID score decreases from 26.0 to 22.7 when $K$ increases from 1 to 2. Additionally, the results suggest that increasing the number of points beyond 2 yields diminishing returns. For instance, on ImageNet with 9 Para. NFE, the FID scores for $K=2$ and $K=3$ are 4.20 and 4.18, respectively, showing minimal improvement.

\noindent\textbf{Latency with different values of $K$.} 
Given that each intermediate gradient is fully parallelizable, we examine whether increasing $K$ noticeably impacts latency. \cref{tab:latency} presents inference latency on a single NVIDIA 4090, evaluated over 1000 generated images with a batch size of 1. We report the average inference time along with the 95\% confidence interval.
For CIFAR-10, FFHQ, and ImageNet, increasing $K$ to 3 does not noticeably impact latency. For LSUN Bedroom, we observe a slight increase in latency when $K=3$. However, earlier results show that $K=2$ already yields significant quality improvements. Therefore, setting $K=2$ provides an effective trade-off, achieving high-quality image generation while avoiding additional inference cost.

\subsection{Ablation Studies}\label{sec:ablations}

\noindent\textbf{Effect of scaling factors.} \cite{ningelucidating,li2024alleviating} identify exposure bias—\ie, the input mismatch between training and sampling—as a key factor leading to error accumulation and sampling drift. To mitigate the bias, they propose scaling the gradient and shifting the timestep. Building on these insights, our \ours~introduces two learnable parameters: $o_n$ and $\delta_n^k$. We compare FID scores without these scaling factors to assess their impact. As shown in ~\cref{tab:scaling}, omitting the scaling factors noticeably reduces image quality. For instance, without $o_n$, FID rises from 4.33 to 5.84 at Para. NFE = 5.

\noindent\textbf{Effect of time schedule.} In ~\cref{tab:schedule}, we present results on CIFAR-10 using commonly used time schedules: LogSNR, EDM, and Time-uniform. Our solver consistently performs better with the time-uniform schedule.

\noindent\textbf{Effect of teacher ODE solvers.}
We study the impact of different teacher ODE solvers in \cref{tab:teacher}. The results show that using DPM-Solver-2 to generate teacher trajectories achieves the best performance.
We hypothesize that this is because DPM-Solver-2 also estimates gradients using intermediate points, resulting in a smaller gap to our \ours.

\begin{figure*}[th]
    \centering
\includegraphics[width=0.95\textwidth]{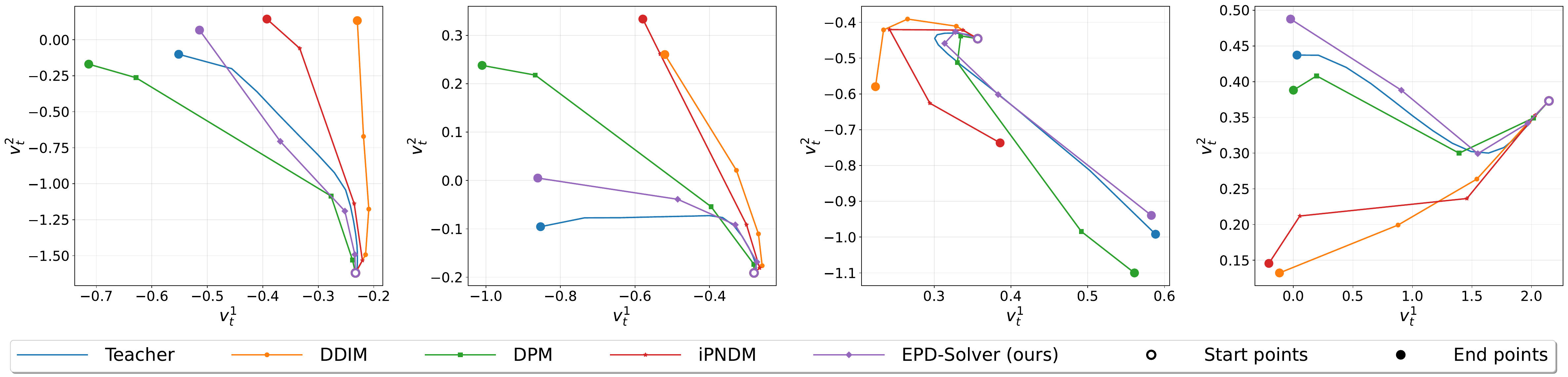}
    \caption{Analysis on local sampling trajectory. The figure shows the generation path of two randomly selected pixels in the images. We employ the  EPD ($\text{Para. NFE}=5, K = 2$) sampler for sampling, using the trajectory of its teacher sampler as the target trajectory. 
     We present the sampling trajectories with $\text{NFE}=5$ of DDIM \cite{songdenoising}, DPM-Solver \cite{lu2022dpm}, and iPNDM \cite{zhangfast} on CIFAR-10 \cite{krizhevsky2009learning}.
    } 
    \label{fig:traj}
\end{figure*}

\begin{figure*}[t!]
    \centering
\includegraphics[width=0.95\textwidth]{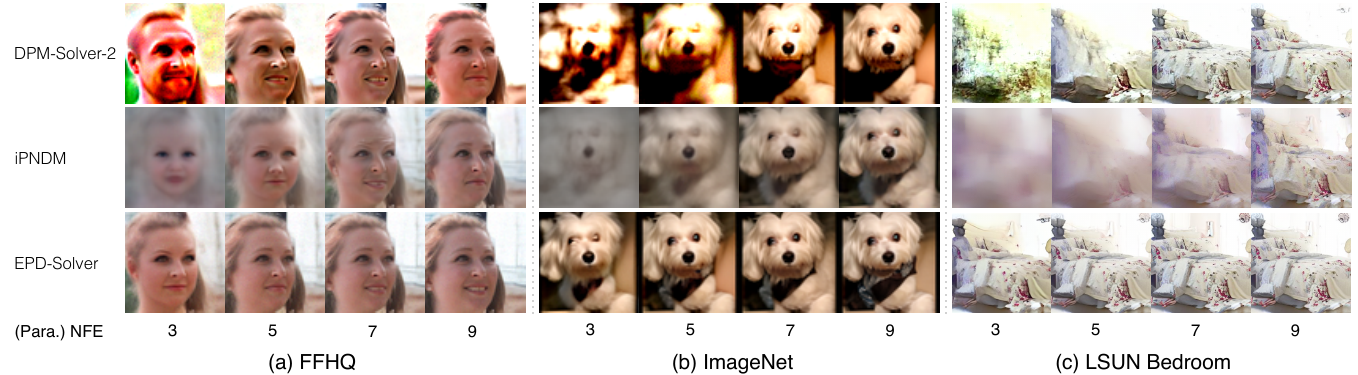}
    \caption{Comparison of generated samples among DPM-Solver-2 \cite{lu2022dpm}, iPNDM \cite{zhangfast} and $\ours$. Compared to other samplers, EPD-Solver achieves high-quality results even at NFE = 3. Additional visualizations are provided in~\cref{sec:visualization}.} 
    \label{Qualitative_results}
\end{figure*}

\subsection{Qualitative Analyses}\label{sec:qualitative}

\noindent\textbf{Qualitative results on trajectory.} Since visualizing the trajectories of high-dimensional data is challenging, we adopt the analysis framework in \cite{liupseudo}. Specifically, as shown in ~\cref{fig:traj}, we randomly select two pixels from the images to perform local trajectory visualization, illustrating how their values evolve during the sampling process. Given the sampling $ \rvx_{t_N}, \rvx_{t_{N-1}}, \dots, \rvx_{t_0} $, we track the corresponding values $ v_{t}^1 $ and $ v_{t}^2 $ at two randomly chosen positions $ p_1 $ and $ p_2 $. We then represent \( (v_{t}^1, v_{t}^2) \) as data points and visualize them in \( \mathbb{R}^2 \). We can clearly observe that the pixel value trajectories of \ours~($\text{Para. NFE}=5 ,K=2$) are closer to the target trajectories compared to other samplers. This shows that our \ours~can generate more accurate trajectory, significantly reducing errors in the sampling process.

\noindent\textbf{Qualitative results on generated samples.} In ~\cref{Qualitative_results}, we compare the generated images from DPM-Solver-2 \cite{lu2022dpm}, iPNDM \cite{zhangfast}, and \ours~using the pretrained models on  FFHQ, ImageNet and LSUN Bedroom. Under the same (Para.) NFE, our \ours~consistently outperforms other samplers in terms of visual perception. This advantage is particularly pronounced in low-NFE settings (NFE = 3, 5), where \ours~is able to generate complete and clear images, while the outputs of other samplers appear highly blurred. These results highlight the superior performance of our method across different NFE settings. Additional visualizations   are provided in~\cref{sec:visualization}.

\section{Conclusion}
\label{sec:conclusion}

In this paper, we propose Ensemble Parallel Direction (EPD), a novel ODE solver that improves diffusion model sampling by leveraging multiple parallel gradient evaluations. Unlike conventional solver-based methods that suffer from truncation errors at low NFE, our approach significantly enhances integral approximation while maintaining low-latency inference. By optimizing a small set of learnable parameters in a distillation fashion, \ours~achieves efficient training and seamless integration into existing diffusion models.
We also generalize our \ours~to \oursplugin, a plugin that can be extended to existing ODE samplers.
Extensive experiments across CIFAR10, FFHQ, ImageNet, LSUN Bedroom, and Stable Diffusion demonstrate that \ours~ consistently outperforms state-of-the-art solvers in FID scores while maintaining computational efficiency. Our findings suggest that parallel gradient estimation is a powerful yet underexplored direction for accelerating diffusion models.

\clearpage

\section*{Acknowledgements}
This research is supported by the RIE2025 Industry Alignment Fund – Industry Collaboration Projects (IAF-ICP) (Award I2301E0026), administered by A*STAR, as well as supported by Alibaba Group and NTU Singapore through Alibaba-NTU Global e-Sustainability CorpLab (ANGEL).

{\small
\bibliographystyle{ieeenat_fullname}
\bibliography{cite}
}

\clearpage
\appendix

\section{Additional Implementation Details}
\subsection{Implementation Details of \ours}\label{sec:implement-epd}
At each sampling step $n$ (from $t_{n+1}$ to $t_n$) in an $N$-step process, the solver provides a set of learned parameters $\Theta_n = \{\tau_n^k, \lambda_n^k, \delta_n^k, o_n\}_{k=1}^K$, implemented as follows:

\noindent \textbf{Intermediate timesteps} ($\tau_n^k$): These are points within $[t_n, t_{n+1}]$, computed via geometric interpolation. Specifically, the interpolation ratio $r_n^k \in [0, 1]$ is obtained by applying a sigmoid to a learnable scalar parameter, yielding
\begin{equation}
    \tau_n^k = t_{n+1}^{r_n^k} \cdot t_n^{1 - r_n^k}.
\end{equation}

\noindent \textbf{Simplex weights} ($\lambda_n^k$): These non-negative weights form a convex combination of the $K$ parallel gradients, satisfying $\sum_{k=1}^K \lambda_n^k = 1$. They are obtained by applying a softmax over $K$ learnable scalar parameters.

\noindent \textbf{Output scaling} ($o_n$): A learnable scalar that scales the overall update direction by a factor of $(1 + o_n)$ to mitigate exposure bias between training and sampling. To implement this, we introduce a per-branch modulation term $\sigma_n^k \in [-0.05, 0.05]$ that scales the corresponding weight $\lambda_n^k$. Specifically, we constrain $\sigma_n^k$ using a sigmoid-based transformation:
\[
\sigma_n^k = 0.1 \times (\texttt{sigmoid}(\tilde{\sigma}_n^k) - 0.5),
\]
where $\tilde{\sigma}_n^k$ is an unconstrained learnable parameter. The final scaling factor is then given by
\[
o_n = \sum_k \lambda_n^k \sigma_n^k - 1.
\]

\noindent \textbf{Timestep shifting} ($\delta_n^k$): A trainable perturbation applied to the intermediate timestep $\tau_n^k$, producing $\tau_n^k + \delta_n^k$ as input to the denoising network. We implement this by introducing a scaling factor $s_n^k$ that transforms $\tau_n^k$ into $s_n^k \tau_n^k$. The relationship between $s_n^k$ and $\delta_n^k$ is given by
\[
s_n^k \tau_n^k = \tau_n^k + \delta_n^k \quad \Rightarrow \quad \delta_n^k = (s_n^k - 1)\tau_n^k.
\]
To prevent overfitting, $s_n^k$ is constrained to a small range (\eg, $[0.95, 1.05]$) using a sigmoid-based transformation. Specifically, we map an unnormalized parameter $\tilde{s}_n^k$ as follows:
\[
s_n^k = 1 + 0.1 \times (\texttt{sigmoid}(\tilde{s}_n^k) - 0.5).
\]

\subsection{Implementation Details of \oursplugin}\label{sec:intro-plugin}
The \oursplugin~serves as a module integrated in any existing ODE solver. We illustrate this using the multi-step iPNDM~\cite{liupseudo,zhangfast} sampler as a representative implementation.
We begin with a brief review of the iPNDM sampler.

\noindent\textbf{Review of iPNDM.} Let $\mathbf{d}_t$ denote the estimated gradient at time step $t$, \ie, $\mathbf{d}_t=\epsilon_\theta(\mathbf{x}_t,t)$. The update at time step $t_n $ is given by:
\begin{align}
    \mathbf{d}_{t_{n+1}}'&=\tfrac{1}{24}(55\mathbf{d}_{t_{n+1}}-59\mathbf{d}_{t_{n+2}}+37\mathbf{d}_{t_{n+3}}-9\mathbf{d}_{t_{n+4}}) \nonumber \\
    \mathbf{x}_{t_n}&=\mathbf{x}_{t_{n+1}}+h_n \mathbf{d}_{t_{n+1}}'.
\end{align}
This rule applies for $n < N - 3$; for brevity, we present only this case. Other cases can be found in the original paper.

\noindent\textbf{Our EPD plugin for iPNDM.} Our plugin replaces $\mathbf{d}_{t_{n+1}}$ with a weighted combination of $K$ parallel intermediate gradients to reduce truncation error. Similar to \ours, we introduce the parameters at step $n$ as $\Theta_n = \{\tau_n^k, \lambda_n^k, \delta_n^k, o_n\}_{k=1}^K$. The gradient  is now estimated as 
\begin{equation}
    \mathbf{d}_{t_{n+1}}^{\mathsf{EPD}}=(1+o_n)\sum_{k=1}^K  \lambda^k_n\bm{\epsilon}_\theta(\rvx_{\tau^k_n},\tau^k_n+\delta^k_n).
\end{equation}
Accordingly, the update for \oursplugin~becomes:
\begin{align}
    \mathbf{d}_{t_{n+1}}'&=\tfrac{1}{24}(55\mathbf{d}_{t_{n+1}}^{\mathsf{EPD}}-59\mathbf{d}_{t_{n+2}}+37\mathbf{d}_{t_{n+3}}-9\mathbf{d}_{t_{n+4}}) \nonumber \\
    \mathbf{x}_{t_n}&=\mathbf{x}_{t_{n+1}}+h_n \mathbf{d}_{t_{n+1}}'.
\end{align}

\oursplugin~incurs minimal training overhead, in line with the lightweight design of the \ours. Thanks to its limited number of learnable parameters, the optimization process is highly efficient. 

\begin{table}[t]
\centering
\fontsize{8}{10}\selectfont
\begin{tabular}{lcccc}
\toprule
\multirow{2}{*}{Timesteps} & \multicolumn{4}{c}{Para. NFE} \\
\cmidrule{2-5} & 3 & 5 & 7 & 9 \\
\midrule
$t_n,t_{n+1}$ (EDM)& 306.2 & 97.67 & 37.28 & 15.76 \\
$\sqrt{t_nt_{n+1}},t_{n+1}$  & 129.6 & 16.51 & 9.86 & 7.06 \\
$\frac{1}{2}(t_n+t_{n+1}),t_{n+1}$ & 105.8 & 36.14 & 18.08 & 9.85 \\
$t_n,\sqrt{t_nt_{n+1}}$  &225.5&130.8&78.49&44.38\\
$t_n, \frac{1}{2}(t_n+t_{n+1})$ &198.6&119.6&59.23&32.21\\ 
$\sqrt{t_nt_{n+1}}, \frac{1}{2}(t_n+t_{n+1})$  &136.1 & 21.17&10.80&5.83 \\
random, $t_{n+1}$ &90.8&30.01&14.37&9.14 \\
random, random &110.7&57.1&22.86& 11.91\\
$\ours, K=2$ & {10.60}  & {5.26}  & {3.29} & {2.52}\\
\bottomrule
\end{tabular}
\caption{FID results on the choices of two intermediate points. Evaluations are conducted on CIFAR-10 \cite{krizhevsky2009learning}. Start point: $t_{n+1}$, end point: $t_n$, midpoints: $\sqrt{t_nt_{n+1}},\frac{1}{2}(t_n+t_{n+1})$, and `random' denotes  a midpoint randomly chosen from $[t_n,t_{n+1}]$.}\label{tab:midpoints_choice}
\end{table}

\subsection{Implementation Details of ParaDiGMS}\label{sec:paradigm_details}

For direct comparison with \texttt{EDP-\{Solver, Plugin\}}, we re-implemented the ParaDiGMS sampler \cite{shih2023parallel} in the EDM \cite{karras2022elucidating} framework, as its public implementation\footnote{https://github.com/AndyShih12/paradigms} is tailored for Stable Diffusion. To ensure a fair latency comparison with our single-GPU \ours, we run ParaDiGMS on two NVIDIA 4090 GPUs, distributing the workload evenly by matching the Para. NFE/GPU ratio.

Specifically, to align the parallel structure with \ours~($K=2$), we set the batch window size of ParaDiGMS to 2. The core principle was to adjust the tolerance parameter, ranging from $1 \times 10^{-2}$ to $1 \times 10^{-1}$, to calibrate the total Para. NFE. The ratio of Para. NFE / GPUs was set to 3, 5, 7 and 9, which ensures the per-GPU workload and latency level for ParaDiGMS roughly matches the single-GPU \ours. We also observed that the efficiency of ParaDiGMS is reduced in low-NFE regimes, as the substantial error per iteration causes its solver stride to frequently set to 1.

\section{Additional Experimental Results}

\begin{table*}[t!]
\small 
\captionsetup[subfloat]{labelformat=simple, labelsep=space}
\begin{minipage}[t]{0.48\textwidth}
    \fontsize{8}{10}\selectfont
     \setlength{\tabcolsep}{4pt} 
    \subfloat[\textbf{CIFAR10} $32 \times 32$ \cite{krizhevsky2009learning}]{
    \centering
        \begin{tabular}{cccccccc}
            \toprule
            Para. NFE &FID& $n$ & $k$ & $r_n^k$ & $s_n^k$ & $\sigma_n^k$ & $\lambda_n^k$ \\
            \midrule
            \multirow{4.5}{*}{3} & \multirow{4.5}{*}{10.40}  
            & \multirow{2}{*}{0} & 0 & 0.01339 & 0.96349 & 0.99731 & 0.85185 \\
            &&  & 1 & 0.67921 & 0.95231 & 0.99754 & 0.14815 \\ \cmidrule{3-8}
            && \multirow{2}{*}{1} & 0 & 0.10020 & 1.03590 & 0.99500 & 0.75008 \\
            &&  & 1 & 0.28855 & 0.95457 & 1.02139 & 0.24992 \\
            \midrule
            \multirow{7}{*}{5} & \multirow{7}{*}{4.33} 
            &\multirow{2}{*}{0} & 0 & 0.03333 & 0.95415 & 0.99735 & 0.86941 \\
            && & 1 & 0.79558 & 0.95376 & 0.98616 & 0.13059 \\ \cmidrule{3-8}
            && \multirow{2}{*}{1} & 0 & 0.07587 & 1.04503 & 0.99400 & 0.41741 \\
            &&  & 1 & 0.63244 & 1.04331 & 1.00711 & 0.58259 \\
            \cmidrule{3-8}
            && \multirow{2}{*}{2} & 0 & 0.38699 & 0.95588 & 1.00299 & 0.22410 \\
            &&  & 1 & 0.09434 & 1.01795 & 0.99999 & 0.77590 \\
            \midrule
            \multirow{9.5}{*}{7}& \multirow{9.5}{*}{2.82} 
            & \multirow{2}{*}{0} & 0 & 0.02511 & 0.96016 & 0.99725 & 0.86908 \\
            &&  & 1 & 0.91820 & 0.95206 & 1.01268 & 0.13092 \\ \cmidrule{3-8}
            && \multirow{2}{*}{1} & 0 & 0.27815 & 0.98792 & 0.98996 & 0.80595 \\
            &&  & 1 & 0.81671 & 0.99280 & 1.01571 & 0.19405 \\
            \cmidrule{3-8}
            &&  \multirow{2}{*}{2} & 0 & 0.34431 & 1.03617 & 0.99038 & 0.17049 \\
            &&  & 1 & 0.60552 & 1.03999 & 0.98517 & 0.82951 \\
            \cmidrule{3-8}
            &&  \multirow{2}{*}{3} & 0 & 0.09416 & 1.01655 & 1.00019 & 0.77621 \\
            &&  & 1 & 0.41999 & 0.96088 & 1.00966 & 0.22379 \\
            \midrule
            \multirow{12}{*}{9} & \multirow{12}{*}{2.49} 
            &\multirow{2}{*}{0} & 0 & 0.28390 & 0.96336 & 0.99459 & 0.74143 \\
            &&  & 1 & 0.08408 & 1.01058 & 0.99785 & 0.25857 \\
            \cmidrule{3-8}
            && \multirow{2}{*}{1} & 0 & 0.33981 & 0.97201 & 0.99713 & 0.31062 \\
            &&  & 1 & 0.47617 & 0.98810 & 1.00195 & 0.68938 \\
            \cmidrule{3-8}
            &&  \multirow{2}{*}{2} & 0 & 0.61703 & 1.03201 & 0.99898 & 0.79387 \\
            &&  & 1 & 0.12204 & 1.01552 & 0.98848 & 0.20613 \\
            \cmidrule{3-8}
            && \multirow{2}{*}{3} & 0 & 0.58062 & 1.02698 & 0.99284 & 0.90470 \\
            &&  & 1 & 0.31738 & 1.02504 & 0.98079 & 0.09530 \\
            \cmidrule{3-8}
            && \multirow{2}{*}{4} & 0 & 0.08719 & 0.98858 & 0.99555 & 0.77554 \\
            &&  & 1 & 0.44045 & 0.97831 & 1.02114 & 0.22446 \\
            \bottomrule
        \end{tabular}
        }
\end{minipage}\hfill
\begin{minipage}[t]{0.48\textwidth}
    \fontsize{8}{10}\selectfont
     \setlength{\tabcolsep}{4pt} 
        \centering
    \subfloat[ \textbf{FFHQ} $64 \times 64$ \cite{karras2019style}]{
        \begin{tabular}{cccccccc}
            \toprule
            Para. NFE &FID& $n$ & $k$ & $r_n^k$ & $s_n^k$ & $\sigma_n^k$ & $\lambda_n^k$ \\
            \midrule
            \multirow{4.5}{*}{3} & \multirow{4.5}{*}{21.74}& \multirow{2}{*}{0} & 0 & 0.00472 & 0.95251 & 0.99909 & 0.85527\\
            &&  & 1 & 0.61291 & 0.95212 & 1.00128 & 0.14473 \\
            \cmidrule{3-8}
            && \multirow{2}{*}{1} & 0 & 0.14636  & 1.00077 & 0.99866 & 0.90603 \\
            &&  & 1 & 0.52375 & 1.03973 &  1.00627 & 0.09397 \\
            \midrule
            \multirow{7}{*}{5} & \multirow{7}{*}{7.84} 
            &\multirow{2}{*}{0} & 0 & 0.00761 & 0.95240 & 0.98863 & 0.85668 \\
            &&  & 1 & 0.68196 & 0.95138 & 1.02573 & 0.14332 \\
            \cmidrule{3-8}
            && \multirow{2}{*}{1} & 0 & 0.48364 & 1.04868 & 1.01419 & 0.98053 \\
            &&  & 1 & 0.19897 & 1.03808 & 1.02313 &  0.01947\\
            \cmidrule{3-8}
            && \multirow{2}{*}{2} & 0 & 0.51289 &  1.01520 &  0.99043 & 0.12838 \\
            &&  & 1 & 0.12570 & 0.96696 & 0.99892 & 0.87162 \\
            \midrule
            \multirow{9.5}{*}{7} & \multirow{9.5}{*}{4.81} 
            &\multirow{2}{*}{0} & 0 & 0.00344 & 0.95175 & 0.99173 & 0.89005 \\
            &&  & 1 & 0.90422 & 0.95040 & 1.01825 & 0.10995 \\
            \cmidrule{3-8}
            && \multirow{2}{*}{1} & 0 & 0.61922 & 1.03974 & 0.99767 & 0.62252 \\
            &&  & 1 & 0.06710 & 1.03036 & 1.00397 & 0.37748 \\
            \cmidrule{3-8}
            && \multirow{2}{*}{2} & 0 & 0.36516 & 1.03981 & 1.01085 & 0.49539\\
            & &  & 1 & 0.71102 & 1.03331 & 1.01083 & 0.50461 \\ 
            \cmidrule{3-8}
            && \multirow{2}{*}{3} & 0 & 0.51302 & 0.99448 & 1.02493 & 0.15205 \\
            &&  & 1 & 0.11444 & 0.96889 & 0.99995 & 0.84795 \\
            \midrule
            \multirow{12}{*}{9} & \multirow{12}{*}{3.82}  
            & \multirow{2}{*}{0}& 0 & 0.07802 & 0.95010 & 0.99990 & 0.16419 \\
            &&  & 1 & 0.08710 & 0.95008 & 0.99990 & 0.83581 \\
            \cmidrule{3-8}
            && \multirow{2}{*}{1} & 0 & 0.85788 & 0.99068 & 0.98106 & 0.00087 \\
            &&  & 1 & 0.51685 & 0.99149 & 0.99980 & 0.99913 \\
            \cmidrule{3-8}
            && \multirow{2}{*}{2} & 0 & 0.5361 & 1.01276 & 0.99527 & 0.68458 \\
            &&  & 1 & 0.49629 & 1.01888 & 0.99385 & 0.31542 \\
            \cmidrule{3-8}
            && \multirow{2}{*}{3} & 0 & 0.55543 & 1.00901 & 1.00370 & 0.83477 \\
            &&  & 1 & 0.95208 & 1.01405& 1.00179 & 0.16523 \\
            \cmidrule{3-8}
            && \multirow{2}{*}{4} & 0 & 0.10233 & 0.95959 & 0.99459 & 0.85282  \\
            &&  & 1 & 0.53488 & 1.03980 & 1.04863 & 0.14718 \\
            \bottomrule
        \end{tabular}
        }
\end{minipage}
\caption{Optimized Parameters for \ours~($K=2$) on CIFAR10 and FFHQ.}
\label{tab:optimized_parameters_all_side_by_side}
\end{table*}
\begin{table*}[t!]
\small
\captionsetup[subfloat]{labelformat=simple, labelsep=space}
\begin{minipage}[t]{0.48\textwidth}
    \fontsize{8}{10}\selectfont
         \setlength{\tabcolsep}{4pt} 
        \subfloat[\textbf{ImageNet} $64 \times 64$ \cite{russakovsky2015imagenet}]{
        \centering
        \begin{tabular}{cccccccc}
            \toprule
            Para. NFE &FID& $n$ & $k$ & $r_n^k$ & $s_n^k$ & $\sigma_n^k$ & $\lambda_n^k$ \\
            \midrule
            \multirow{4.5}{*}{3} & \multirow{4.5}{*}{18.28} & \multirow{2}{*}{0} & 0 & 0.03892 & 0.90820 & 0.99810 & 0.78701 \\
            &&  & 1 & 0.58080 & 0.95077 & 1.00097 & 0.21299 \\ 
            \cmidrule{3-8}
            && \multirow{2}{*}{1} & 0 & 0.18326 & 0.99336 & 0.99910 & 0.97757 \\
            &&  & 1 & 0.08246 & 1.01142 & 1.02640 & 0.02243 \\
            \midrule
            \multirow{7}{*}{5}  & \multirow{7}{*}{6.35} & \multirow{2}{*}{0} & 0 &  0.14336 & 0.90835 & 0.99266 & 0.78550 \\
            &&  & 1 & 0.54204 & 0.93916 & 0.99114 &  0.21450 \\
            \cmidrule{3-8}
            && \multirow{2}{*}{1} & 0 & 0.71830 & 1.08078 & 1.00955 & 0.49788 \\
            &&  & 1 & 0.39094 & 1.07179 & 1.01071 & 0.50212 \\
            \cmidrule{3-8}
            && \multirow{2}{*}{2} & 0 & 0.25820 & 0.96964 & 1.00597 & 0.37857 \\
            &&  & 1 & 0.10124 & 1.00380 & 1.00316 & 0.62143 \\
            \midrule
            \multirow{9.5}{*}{7} & \multirow{9.5}{*}{5.26} & \multirow{2}{*}{0} & 0 & 0.11952 & 0.90686 & 0.99347 & 0.91217 \\
            &&  & 1 & 0.95726 & 0.91100 & 1.01887 & 0.08783 \\
            \cmidrule{3-8}
            && \multirow{2}{*}{1} & 0 & 0.41813 & 1.03421 & 0.99877 & 0.83649 \\
            &&  & 1 & 0.76716 & 1.04605 & 1.00396 & 0.16351 \\
            \cmidrule{3-8}
            && \multirow{2}{*}{2} & 0 & 0.86120 & 1.03538 & 1.00931 & 0.02866 \\
            &&  & 1 & 0.52961 & 1.04485 & 1.00040 & 0.97134 \\
            \cmidrule{3-8}
            && \multirow{2}{*}{3} & 0 & 0.19129 & 0.98157 & 1.0024 & 0.99873 \\
            &&  & 1 & 0.17888 & 0.99072 & 1.02263 & 0.00127 \\
            \midrule
            \multirow{12}{*}{9} & \multirow{12}{*}{4.27} & \multirow{2}{*}{0} & 0 & 0.97878 & 0.90410 & 1.01060 & 0.04239 \\
            &&  & 1 & 0.12206 & 0.90047 & 0.99891 & 0.95761 \\
            \cmidrule{3-8}
            && \multirow{2}{*}{1} & 0 & 0.40113 & 0.97924 & 0.99857 & 0.90324 \\
            &&  & 1 & 0.84037 & 1.04647 & 0.99850 & 0.09676 \\
            \cmidrule{3-8}
            && \multirow{2}{*}{2} & 0 & 0.55210 & 1.00744 & 0.99590 & 0.99983 \\
            &&  & 1 & 0.17699 & 0.97798 & 1.01484 & 0.00017 \\
            \cmidrule{3-8}
            && \multirow{2}{*}{3} & 0 & 0.67823 & 0.99619 & 1.01995 & 0.99919 \\
            &&  & 1 & 0.89296 & 1.02559 & 1.02289 & 0.00081 \\
            \cmidrule{3-8}
            && \multirow{2}{*}{4} & 0 & 0.26663 & 0.91395 & 1.01391 & 0.60252 \\
            &&  & 1 &  0.00584 & 1.06452 & 1.00333 & 0.39748 \\
            \bottomrule
        \end{tabular}
        }
\end{minipage}\hfill
\begin{minipage}[t]{0.48\textwidth}
    \fontsize{8}{10}\selectfont
     \setlength{\tabcolsep}{4pt} 
    \centering
        \subfloat[ \textbf{LSUN Bedroom} $256 \times 256$~\cite{yu2015lsun}]{
        \centering
        \begin{tabular}{cccccccc}
            \toprule
            Para. NFE &FID& $n$ & $k$ & $r_n^k$ & $s_n^k$ & $\sigma_n^k$ & $\lambda_n^k$ \\
            \midrule
            \multirow{4.5}{*}{3} & \multirow{4.5}{*}{13.21} & \multirow{2}{*}{0} & 0 & 0.82995 & 0.98769 & 1.01204 & 0.09938 \\
            &&  & 1 & 0.0410 & 1.0101 & 0.9989 & 0.9006 \\
            \cmidrule{3-8}
            && \multirow{2}{*}{1} & 0 & 0.03654 & 1.00350 & 0.98716 & 0.01419 \\
            &&  & 1 & 0.22279 & 0.97061 & 1.00927 & 0.98581 \\
            \midrule
            \multirow{7}{*}{5} & \multirow{7}{*}{7.52} & \multirow{2}{*}{0} & 0 & 0.99712 & 1.00000 & 0.99752 & 0.07831 \\
            &&  & 1 & 0.02895 & 1.00000 & 1.00046 & 0.92169 \\
            \cmidrule{3-8}
            && \multirow{2}{*}{1} & 0 & 0.52144 & 1.00000 & 1.00186 & 0.61657 \\
            &&  & 1 & 0.18287 & 1.00000 & 0.99460 & 0.38343 \\
            \cmidrule{3-8}
            && \multirow{2}{*}{2} & 0 & 0.20350 & 1.00000 & 0.96961 & 0.24707 \\
            &&  & 1 & 0.23099 & 1.00000 & 1.00159 & 0.75293 \\
            \midrule
            \multirow{9.5}{*}{7} & \multirow{9.5}{*}{5.97} & \multirow{2}{*}{0} & 0 & 0.92247 & 1.00000 & 1.00783 & 0.00004 \\
            &&  & 1 & 0.02283 & 1.00000 & 0.99966 & 1.00000 \\
            \cmidrule{3-8}
            && \multirow{2}{*}{1} & 0 & 0.45881 & 1.00000 & 1.00193 & 0.46663 \\
            &&  & 1 & 0.54699 & 1.00000 & 1.00185 & 0.53337 \\
            \cmidrule{3-8}
            && \multirow{2}{*}{2} & 0 & 0.09864 & 1.00000 & 0.98422 & 0.06541 \\
            &&  & 1 & 0.46885 & 1.00000 & 0.99675 & 0.93459 \\
            \cmidrule{3-8}
            && \multirow{2}{*}{3} & 0 & 0.20864 & 1.00000 & 0.96134 & 0.98301 \\
            &&  & 1 & 0.09425 & 1.00000 & 1.02840 & 0.01699 \\
            \midrule
            \multirow{12}{*}{9} & \multirow{12}{*}{5.01}& \multirow{2}{*}{0} & 0 & 0.87854 & 1.00000 & 1.00569 & 0.07317 \\
            &&  & 1 & 0.07964 & 1.00000 & 0.99953 & 0.92683 \\
            \cmidrule{3-8}
            && \multirow{2}{*}{1} & 0 & 0.40848 & 1.00000 & 0.99842 & 0.82916 \\
            &&  & 1 & 0.94301 & 1.00000 & 1.00355 & 0.17084 \\
            \cmidrule{3-8}
            && \multirow{2}{*}{2} & 0 & 0.67654 & 1.00000 & 1.00375 & 0.01636 \\
            &&  & 1 & 0.49911 & 1.00000 & 1.00348 & 0.98364 \\
            \cmidrule{3-8}
            && \multirow{2}{*}{3} & 0 & 0.45169 & 1.00000 & 0.98647 & 0.14504 \\
            &&  & 1 & 0.40655 & 1.00000 & 0.99226 & 0.85496 \\
            \cmidrule{3-8}
            && \multirow{2}{*}{4} & 0 & 0.30053 & 1.00000 & 1.00438 & 0.02853 \\
            &&  & 1 & 0.20058 & 1.00000 & 0.95733 & 0.97147 \\
            \bottomrule
        \end{tabular}
        }
\end{minipage}
\caption{Optimized Parameters for \ours~($K=2$) on ImageNet and LSUN Bedroom.}
\label{tab:optimized_parameters_imagenet_lsun_side}
\end{table*}

\noindent\textbf{Other choice of intermediate points.}
In ~\cref{tab:midpoints_choice}, we compare our $\ours$ with $K=2$, \ie, two learned intermediate points, against two manually selected midpoints and randomly selected ones. In particular, the manually selected midpoints include the start timestep $t_n$, the end timestep $t_{n+1}$ (adopted in EDM), the geometric mean $\sqrt{t_n t_{n+1}}$ (used in DPM-Solver-2), and the arithmetic mean $\frac{1}{2}(t_n+t_{n+1})$. The random midpoints are uniformly sampled from $[t_n, t_{n+1}]$. We note several observations: (1) The combination of start points with mean points (geometric and arithmetic) significantly outperforms combinations that include the end point. For example, using the geometric and arithmetic points achieves an FID of 5.83 with NFE = 9, whereas incorporating the end point leads to much higher FID scores — 44.38 and 32.21 for the geometric and arithmetic points, respectively. (2) The combination that includes random points achieves competitive results. For instance, using a random point together with the start point yields better FID scores than EDM across all NFE values. (3) The gap between the best combination of handcrafted intermediate timesteps and our learned ones remains large, highlighting the necessity of our proposed method.

\subsection{Optimized Parameters for \ours}\label{sec:learnedparameters}

We provide our optimized parameters of \ours~with $K=2$ for CIFAR-10, ImageNet, FFHQ and LSUN Bedroom in \cref{tab:optimized_parameters_all_side_by_side,tab:optimized_parameters_imagenet_lsun_side} with different Para. NFEs. 
According to the implementation details in~\cref{sec:implement-epd}, the parameters $\tau_n^k ,\delta_n^k,o_n$ are derived as follows:
\begin{align}
    \tau_n^k&=t_{n+1}^{r_n^k} \cdot t_n^{1 - r_n^k} \\
    \delta_n^k&=(s_n^k-1)\tau_n^k \\ 
    o_n&=\sum_k \lambda_n^k\sigma_n^k - 1
\end{align}

\begin{table*}[t!]
\small 
\captionsetup[subfloat]{labelformat=simple, labelsep=space}
\begin{minipage}[t]{0.48\textwidth}
    \fontsize{8}{10}\selectfont
     \setlength{\tabcolsep}{4pt} 
    \subfloat[\textbf{CIFAR10} $32 \times 32$ \cite{krizhevsky2009learning}]{
    \centering
        \begin{tabular}{cccccccc}
            \toprule
            Para. NFE &FID& $n$ & $k$ & $r_n^k$ & $s_n^k$ & $\sigma_n^k$ & $\lambda_n^k$ \\
            \midrule
            \multirow{4.5}{*}{3} & \multirow{4.5}{*}{10.54}  
            & \multirow{2}{*}{0} & 0 & 0.06837 & 0.81145 &0.99957  & 0.91271 \\
            &&  & 1 & 0.68803 & 0.85836 & 0.99981 & 0.08729 \\ \cmidrule{3-8}
            && \multirow{2}{*}{1} & 0 & 0.12320 & 0.97533 & 0.99903 & 0.85072 \\
            &&  & 1 & 0.28206 & 0.85043 & 1.00671 & 0.14928 \\
            \midrule
            \multirow{7}{*}{5} & \multirow{7}{*}{4.47} 
            &\multirow{2}{*}{0} & 0 & 0.10548 & 0.80808 & 0.99606 & 0.95656 \\
            && & 1 & 0.96750 & 0.89210 & 1.00082 & 0.04344 \\ \cmidrule{3-8}
            && \multirow{2}{*}{1} & 0 & 0.04114 & 1.03816 & 1.00480 & 0.52907 \\
            &&  & 1 & 0.57891 & 1.02063 & 1.02490 & 0.47093 \\
            \cmidrule{3-8}
            && \multirow{2}{*}{2} & 0 & 0.27989 & 1.00150 & 0.95600 & 0.26331 \\
            &&  & 1 & 0.05394 & 1.02182 & 0.98523 & 0.73669 \\
            \midrule
            \multirow{9.5}{*}{7}& \multirow{9.5}{*}{3.27} 
            & \multirow{2}{*}{0} & 0 & 0.08991 & 0.80504 & 0.99845 & 0.94689 \\
            &&  & 1 & 0.94988 & 0.95487 & 1.01496 & 0.05311 \\ \cmidrule{3-8}
            && \multirow{2}{*}{1} & 0 & 0.04569 & 0.88770 & 0.99774 & 0.75623 \\
            &&  & 1 &  0.80305& 1.04391 & 0.99378 & 0.24377 \\
            \cmidrule{3-8}
            &&  \multirow{2}{*}{2} & 0 & 0.91959 & 1.10578 & 0.99989 & 0.00408 \\
            &&  & 1 & 0.42678 & 1.01745 & 1.00242 & 0.99592 \\
            \cmidrule{3-8}
            &&  \multirow{2}{*}{3} & 0 & 0.36480 & 0.90472 & 1.02327 &0.20787  \\
            &&  & 1 & 0.07649 & 0.96814 & 1.00433 & 0.79213 \\
            \midrule
            \multirow{12}{*}{9} & \multirow{12}{*}{2.42} 
            &\multirow{2}{*}{0} & 0 & 0.08244 & 0.80210 & 0.99483 & 0.08638 \\
            &&  & 1 & 0.25440 & 0.81528 & 0.99964 & 0.91362 \\
            \cmidrule{3-8}
            && \multirow{2}{*}{1} & 0 & 0.02193 & 0.80719 & 0.99517 & 0.99163 \\
            &&  & 1 & 0.02935  & 0.88719 & 0.99437 & 0.00837 \\
            \cmidrule{3-8}
            &&  \multirow{2}{*}{2} & 0 & 0.25227 & 1.08671 & 0.99438 & 0.02010 \\
            &&  & 1 & 0.55490 & 1.03722 & 0.99923 & 0.97990 \\
            \cmidrule{3-8}
            && \multirow{2}{*}{3} & 0 & 0.48861 & 1.01472 & 1.00312 & 0.81266 \\
            &&  & 1 & 0.02553 & 0.98693 &  1.00521& 0.18734 \\
            \cmidrule{3-8}
            && \multirow{2}{*}{4} & 0 & 0.07257 & 0.97384 & 0.99552 & 0.78925 \\
            &&  & 1 & 0.39513  & 0.96933 & 0.99003 & 0.21075 \\
            \bottomrule
        \end{tabular}
        }
\end{minipage}\hfill
\begin{minipage}[t]{0.48\textwidth}
    \fontsize{8}{10}\selectfont
     \setlength{\tabcolsep}{4pt} 
        \centering
    \subfloat[ \textbf{FFHQ} $64 \times 64$ \cite{karras2019style}]{
        \begin{tabular}{cccccccc}
            \toprule
            Para. NFE &FID& $n$ & $k$ & $r_n^k$ & $s_n^k$ & $\sigma_n^k$ & $\lambda_n^k$ \\
            \midrule
            \multirow{4.5}{*}{3} &\multirow{4.5}{*}{19.02}& \multirow{2}{*}{0} & 0 & 0.07642 & 0.84410 & 0.99934 & 0.94986 \\
            &&  & 1 & 0.91510 & 0.97713 & 1.01079 & 0.05014 \\
            \cmidrule{3-8}
            && \multirow{2}{*}{1} & 0 & 0.17864 & 0.97337 & 1.00023 & 0.99041 \\
            &&  & 1 & 0.15293 & 0.90787 & 1.02719 & 0.00959 \\
            \midrule
            \multirow{7}{*}{5} & \multirow{7}{*}{7.97}  &\multirow{2}{*}{0} & 0 & 0.00858 & 0.82007 & 0.99986 & 0.87461 \\
            &&  & 1 & 0.65658 & 0.86946 & 0.99954 & 0.12539 \\
            \cmidrule{3-8}
            && \multirow{2}{*}{1} & 0 & 0.39945 & 0.99765 & 1.00157 & 0.99812 \\
            &&  & 1 & 0.18867 & 1.03054 & 1.01357 & 0.00188 \\
            \cmidrule{3-8}
            && \multirow{2}{*}{2} & 0 & 0.33148 & 0.96555 & 0.99766 & 0.22642 \\
            &&  & 1 & 0.07594 & 0.97690 & 0.99730 & 0.77358 \\
            \midrule
            \multirow{9.5}{*}{7} &\multirow{9.5}{*}{5.09}  &\multirow{2}{*}{0} & 0 & 0.01069 & 0.81532 & 0.99965 & 0.92015 \\
            &&  & 1 & 0.85634 & 0.86078 & 0.99965 & 0.07985 \\
            \cmidrule{3-8}
            && \multirow{2}{*}{1} & 0 & 0.37517 & 1.00369 & 0.99838 & 0.88685 \\
            &&  & 1 & 0.71151 & 1.00119 & 1.00481 & 0.11315 \\
            \cmidrule{3-8}
            && \multirow{2}{*}{2} & 0 & 0.08475 & 1.04325 & 1.03287 & 0.00052 \\
            & &  & 1 & 0.38954 & 1.00524 & 1.00463 & 0.99948 \\ 
            \cmidrule{3-8}
            && \multirow{2}{*}{3} & 0 & 0.08461 & 0.98373 & 0.98399 & 0.76003 \\
            &&  & 1 & 0.39386 & 1.01515 & 0.97975 & 0.23997 \\
            \midrule
            \multirow{12}{*}{9} & \multirow{12}{*}{3.53} &\multirow{2}{*}{0} & 0 & 0.94960 & 0.82963 & 1.00126 & 0.06572 \\
            &&  & 1 & 0.00362 & 0.82194 & 0.9998 & 0.93428 \\
            \cmidrule{3-8}
            && \multirow{2}{*}{1} & 0 & 0.06822 & 0.87369 & 0.99903 & 0.19003 \\
            &&  & 1 & 0.48656 & 1.01113 & 0.99772 & 0.80995 \\
            \cmidrule{3-8}
            && \multirow{2}{*}{2} & 0 & 0.38262 & 1.02269 & 0.99920 & 0.84123 \\
            &&  & 1 & 0.98681 & 0.99794 & 1.01047 & 0.15877 \\
            \cmidrule{3-8}
            && \multirow{2}{*}{3} & 0 & 0.08146 & 0.99005 & 1.01881 & 0.56715 \\
            &&  & 1 & 0.89689 & 1.01201 & 0.99138 & 0.43285 \\
            \cmidrule{3-8}
            && \multirow{2}{*}{4} & 0 & 0.07455 & 0.96557 & 0.97884 & 0.80133 \\
            &&  & 1 & 0.47558 & 1.09918 & 0.95222 & 0.19867 \\
            \bottomrule
        \end{tabular}
        }
\end{minipage}
\caption{Optimized Parameters for \oursplugin~($K=2$) on CIFAR10 and FFHQ.}
\label{tab:epd-plugin-cifar-ffhq}
\end{table*}
\begin{table*}[t!]
\small
\captionsetup[subfloat]{labelformat=simple, labelsep=space}
\begin{minipage}[t]{0.48\textwidth}
    \fontsize{8}{10}\selectfont
         \setlength{\tabcolsep}{4pt} 
        \subfloat[\textbf{ImageNet} $64 \times 64$ \cite{russakovsky2015imagenet}]{
        \centering
        \begin{tabular}{cccccccc}
            \toprule
            Para. NFE &FID& $n$ & $k$ & $r_n^k$ & $s_n^k$ & $\sigma_n^k$ & $\lambda_n^k$ \\
            \midrule
            \multirow{4.5}{*}{3} & \multirow{4.5}{*}{19.89} 
            & \multirow{2}{*}{0} & 0 & 0.01805  & 0.89265  & 0.99984 & 0.81070 \\
            &&  & 1 & 0.59732  & 0.95910 & 0.99862  & 0.18930\\ 
            \cmidrule{3-8}
            && \multirow{2}{*}{1} & 0 & 0.15989 &0.96659  &1.00771  & 0.96197\\
            &&  & 1 &  0.26658 & 0.89747 & 1.04079 & 0.03803\\
            \midrule
            \multirow{7}{*}{5}  & \multirow{7}{*}{8.17} 
            & \multirow{2}{*}{0} & 0 & 0.11246 & 0.82261 & 0.99876 & 0.92199\\
            &&  & 1 & 0.92205  & 0.96191 & 1.01100 & 0.07801 \\
            \cmidrule{3-8}
            && \multirow{2}{*}{1} & 0 & 0.00511  &0.97233  &0.99878 & 0.45635\\
            &&  & 1 & 0.61007 &  0.99912  & 1.00419 & 0.54365\\
            \cmidrule{3-8}
            && \multirow{2}{*}{2} & 0 & 0.35416 & 0.92432 & 0.99057 & 0.04391\\
            &&  & 1 & 0.13234  &  0.96354  &  0.99885 & 0.95609  \\
            \midrule
            \multirow{9.5}{*}{7} & \multirow{9.5}{*}{4.81} 
            & \multirow{2}{*}{0} & 0 & 0.14306  & 0.82532 &0.99963  & 0.99640 \\
            &&  & 1 & 0.02764 & 0.94802  & 0.96580& 0.00360\\
            \cmidrule{3-8}
            && \multirow{2}{*}{1} & 0 & 0.46578 & 0.98602 &1.00224 & 0.99615\\
            &&  & 1 &  0.09086 &  1.08617  &1.02104 & 0.00385\\
            \cmidrule{3-8}
            && \multirow{2}{*}{2} & 0 & 0.04504 & 1.05987 &  1.01408 & 0.00020\\
            &&  & 1  & 0.44154 &  0.99292  &  0.99536 & 0.99980 \\
            \cmidrule{3-8}
            && \multirow{2}{*}{3} & 0 & 0.03175 & 0.90298 & 0.98815  &  0.00276\\
            &&  & 1 & 0.14969 &  0.94543 &1.00853 &  0.99724 \\
            \midrule
            \multirow{12}{*}{9} & \multirow{12}{*}{4.02} & \multirow{2}{*}{0} 
            & 0 & 0.33263 & 0.84332 & 0.99983 & 0.12259\\
            &&  & 1 &  0.13371  & 0.85792 & 0.99931  & 0.87741\\
            \cmidrule{3-8}
            && \multirow{2}{*}{1} & 0&  0.05410 & 0.89662 & 1.00055 &0.24089  \\
            &&  & 1 & 0.54876 & 0.99484  &  0.99886  & 0.75911 \\
            \cmidrule{3-8}
            && \multirow{2}{*}{2} & 0 & 0.37444 & 1.00578  &1.00105 & 0.88450\\
            &&  & 1 &  0.94384 &  1.01652  &0.98910 & 0.11550\\
            \cmidrule{3-8}
            && \multirow{2}{*}{3} & 0 & 0.28771 & 1.00243 &  0.99434 & 0.76097  \\
            &&  & 1 &  0.82883 & 1.00291 &  0.99311 & 0.23903\\
            \cmidrule{3-8}
            && \multirow{2}{*}{4} & 0 &0.11117  & 0.98196 & 1.01350& 0.80293\\
            &&  & 1 & 0.41243 & 0.88880 & 1.08111&  0.19707  \\
            \bottomrule
        \end{tabular}
        }
\end{minipage}\hfill
\begin{minipage}[t]{0.48\textwidth}
    \fontsize{8}{10}\selectfont
     \setlength{\tabcolsep}{4pt} 
    \centering
        \subfloat[ \textbf{LSUN Bedroom} $256 \times 256$~\cite{yu2015lsun}]{
        \centering
        \begin{tabular}{cccccccc}
            \toprule
            Para. NFE &FID& $n$ & $k$ & $r_n^k$ & $s_n^k$ & $\sigma_n^k$ & $\lambda_n^k$ \\
            \midrule
            \multirow{4.5}{*}{3} & \multirow{4.5}{*}{14.12} 
            & \multirow{2}{*}{0} & 0 & 0.78697 & 1.00000 & 1.00375 & 0.10230 \\
            &&  & 1 & 0.02085 & 1.00000 & 0.99945  & 0.89770\\ 
            \cmidrule{3-8}
            && \multirow{2}{*}{1} & 0 & 0.08334 & 1.00000 & 0.96782 &0.18352  \\
            &&  & 1 & 0.23899 &  1.00000 & 0.99524 & 0.81648\\
            \midrule
            \multirow{7}{*}{5}  & \multirow{7}{*}{8.26} 
            & \multirow{2}{*}{0} & 0 & 0.97220  & 0.98923 &1.00016  & 0.07808\\
            &&  & 1 & 0.03306 & 1.00415 & 0.99991 &0.92192  \\
            \cmidrule{3-8}
            && \multirow{2}{*}{1} & 0 & 0.52337 & 0.99607 & 1.00463 & 0.60203 \\
            &&  & 1 & 0.01602   & 1.00079 & 0.99249 &0.39797   \\
            \cmidrule{3-8}
            && \multirow{2}{*}{2} & 0 &  0.12524 & 0.99813  &  0.96174 & 0.49642  \\
            &&  & 1 & 0.29699 & 0.99950  & 1.01130  & 0.50358 \\
            \midrule
            \multirow{9.5}{*}{7} & \multirow{9.5}{*}{5.24} 
            & \multirow{2}{*}{0} & 0 & 0.97094 & 0.98527 & 1.01234 & 0.06101 \\
            &&  & 1 & 0.07156 & 1.00461 & 0.99893 & 0.93899\\
            \cmidrule{3-8}
            && \multirow{2}{*}{1} & 0 & 0.70513 & 0.99016 & 1.01166 & 0.32484 \\
            &&  & 1 & 0.24738 &  0.98946 & 0.99696 & 0.67516 \\
            \cmidrule{3-8}
            && \multirow{2}{*}{2} & 0 &  0.27565 & 1.01344 & 0.97876 & 0.57267 \\
            &&  & 1  &  0.54473 & 1.00123 & 1.00931 & 0.42733 \\
            \cmidrule{3-8}
            && \multirow{2}{*}{3} & 0 & 0.16616 & 0.98549 & 0.96569 & 0.85584 \\
            &&  & 1 & 0.38606 & 0.99734 & 1.02813 & 0.14416 \\
            \midrule
            \multirow{12}{*}{9} & \multirow{12}{*}{4.51} & \multirow{2}{*}{0} 
            & 0 & 0.17020 & 1.01750 & 0.99792 & 0.34563\\
            &&  & 1 & 0.01271 & 0.99479 &1.00060 & 0.65437 \\
            \cmidrule{3-8}
            && \multirow{2}{*}{1} & 0& 0.43953 & 0.98534 & 0.99969 & 0.96036\\
            &&  & 1 & 0.82230 & 0.99246  &0.99977 & 0.03964\\
            \cmidrule{3-8}
            && \multirow{2}{*}{2} & 0 &  0.25682 & 1.00056 &1.00433 &0.30549 \\
            &&  & 1 & 0.50732 & 1.00773 & 0.99838 & 0.69451\\
            \cmidrule{3-8}
            && \multirow{2}{*}{3} & 0 & 0.29627 & 1.01221 & 0.98564 &  0.31065  \\
            &&  & 1 & 0.48616 &  1.01091 & 0.99254 & 0.68935 \\
            \cmidrule{3-8}
            && \multirow{2}{*}{4} & 0 & 0.32949 & 1.00615 &0.98884 & 0.04682\\
            &&  & 1 & 0.19802 &  0.98760 & 0.95685 & 0.95318 \\
            \bottomrule
        \end{tabular}
        }
\end{minipage}
\caption{Optimized Parameters for \oursplugin~($K=2$) on ImageNet and LSUN Bedroom.}
\label{tab:optimized_parameters_imagenet_lsun_plugin}
\end{table*}
\subsection{Optimized Parameters for \oursplugin}
We provide our optimized parameters of \oursplugin~with $K=2$ for CIFAR10, ImageNet, FFHQ and LSUN Bedroom in~\cref{tab:epd-plugin-cifar-ffhq,tab:optimized_parameters_imagenet_lsun_plugin} with different Para.NFEs. 

\subsection{Additional Qualitative Results}\label{sec:visualization}
\begin{figure*}[t]
\centering
\includegraphics[width=.95\textwidth]{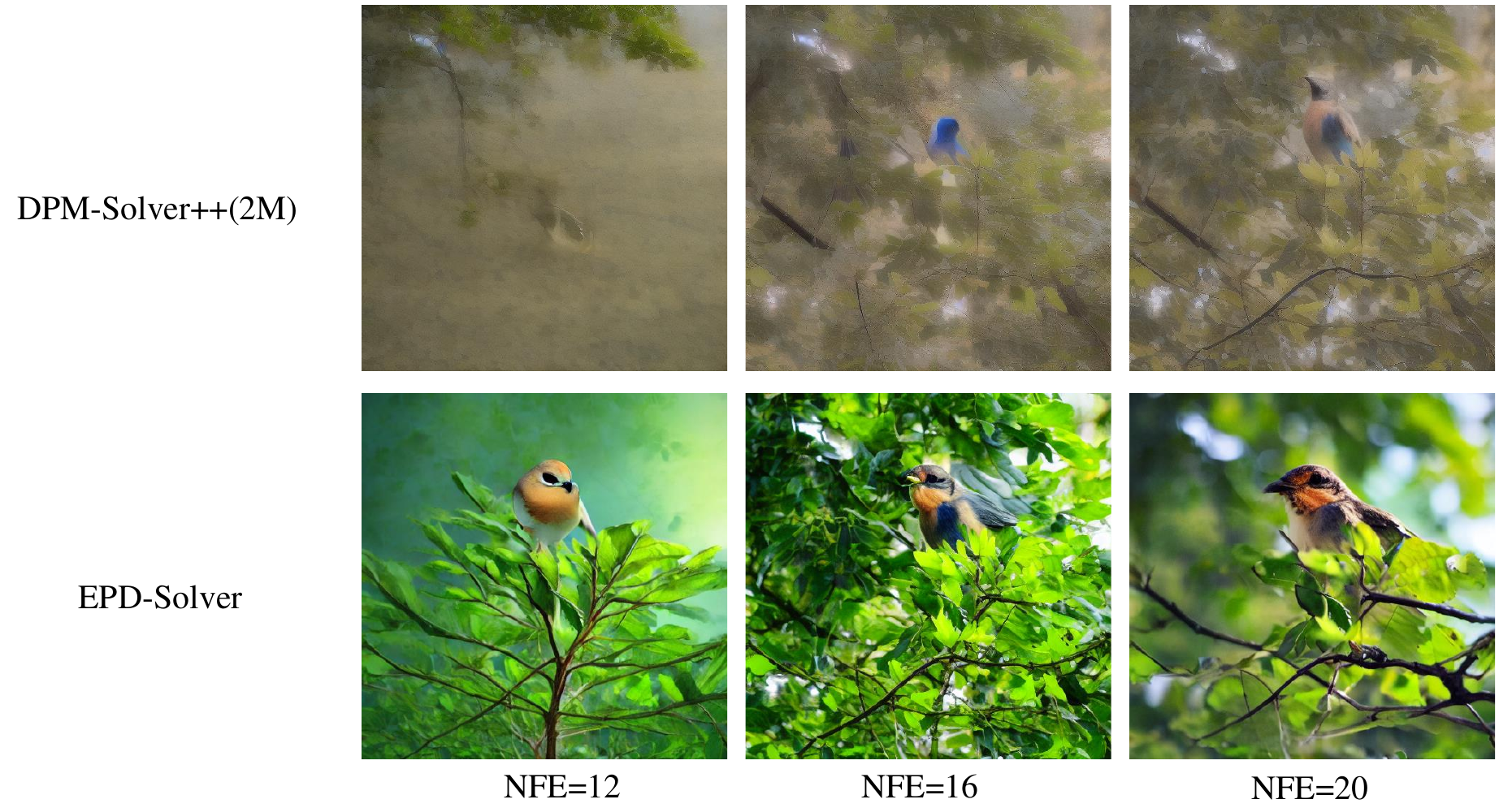}
\caption{Comparison of image generation quality between DPM-Solver++ (2M) and \ours at different (Para.) NFEs.}
\label{fig:method}
\end{figure*}

Here, we show some qualitative results on different datasets in \cref{fig:method,fig:sup_grid_cifar10_3,fig:sup_grid_ffhq_3,fig:sup_grid_img_3}.

\clearpage


\newcommand{\nfea}{\text{3}}
\newcommand{\nfeb}{\text{9}}

\begin{figure*}[t]
  \centering
  \begin{subfigure}[b]{0.48\linewidth}
      \includegraphics[width=\linewidth]{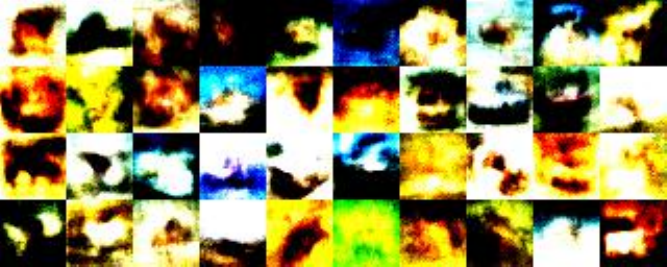}
      \caption{DPM-Solver-2. NFE=\(\nfea\)}
  \end{subfigure}
  \begin{subfigure}[b]{0.48\linewidth}
      \includegraphics[width=\linewidth]{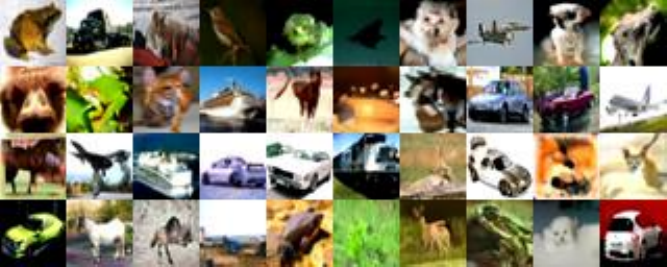}
      \caption{DPM-Solver-2. NFE=\(\nfeb\)}
  \end{subfigure}
  \begin{subfigure}[b]{0.48\linewidth}
      \includegraphics[width=\linewidth]{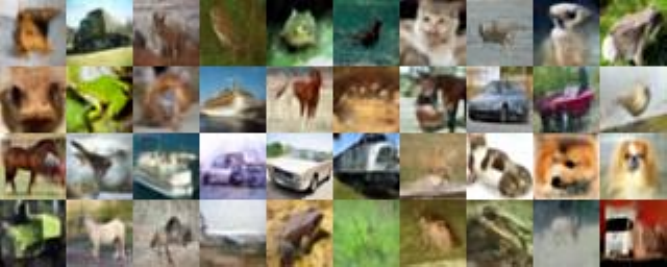}
      \caption{\(\ours\). Para. NFE=\(\nfea\)}
  \end{subfigure}
  \begin{subfigure}[b]{0.48\linewidth}
      \includegraphics[width=\linewidth]{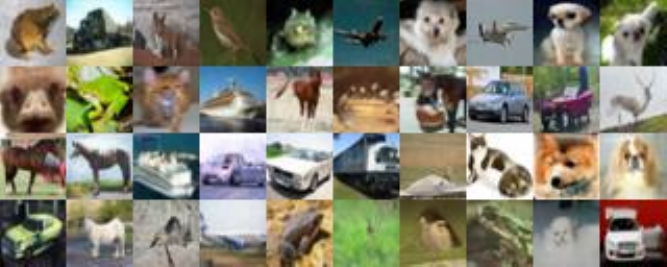}
      \caption{\(\ours\). Para. NFE=\(\nfeb\)}
  \end{subfigure}
  \caption{Qualitative result on CIFAR10 32$\times$32 (\(\nfea\) and \(\nfeb\) NFEs)}
\label{fig:sup_grid_cifar10_3}
\end{figure*}

\begin{figure*}[t]
  \centering
  \begin{subfigure}[b]{0.48\linewidth}
      \includegraphics[width=\linewidth]{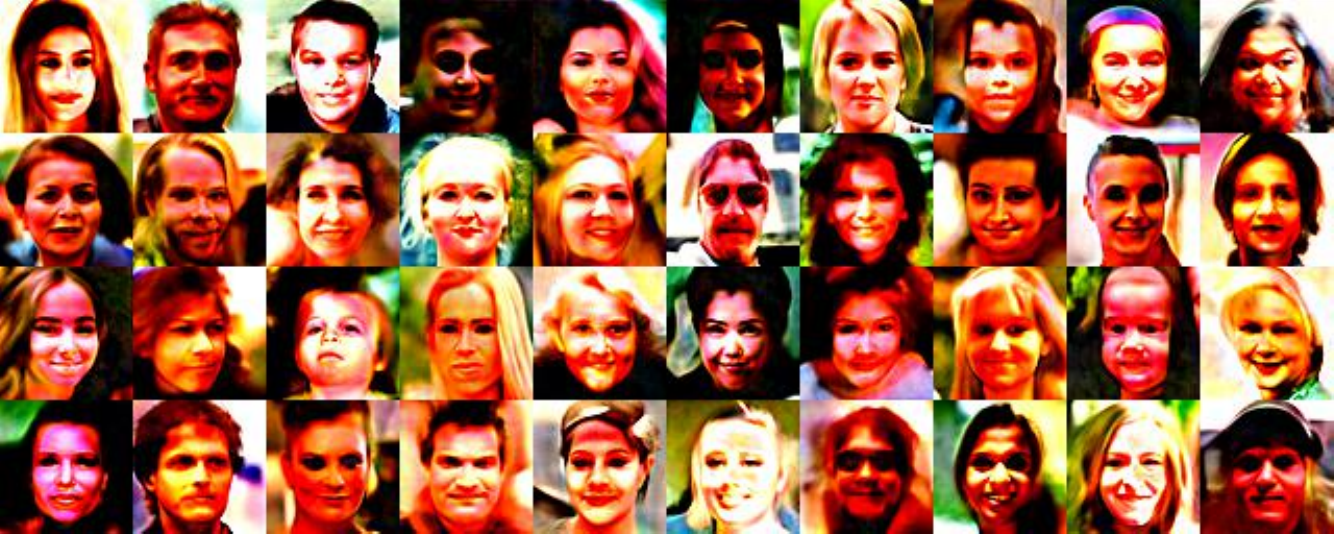}
      \caption{DPM-Solver-2. NFE=\(\nfea\)}
  \end{subfigure}
  \begin{subfigure}[b]{0.48\linewidth}
      \includegraphics[width=\linewidth]{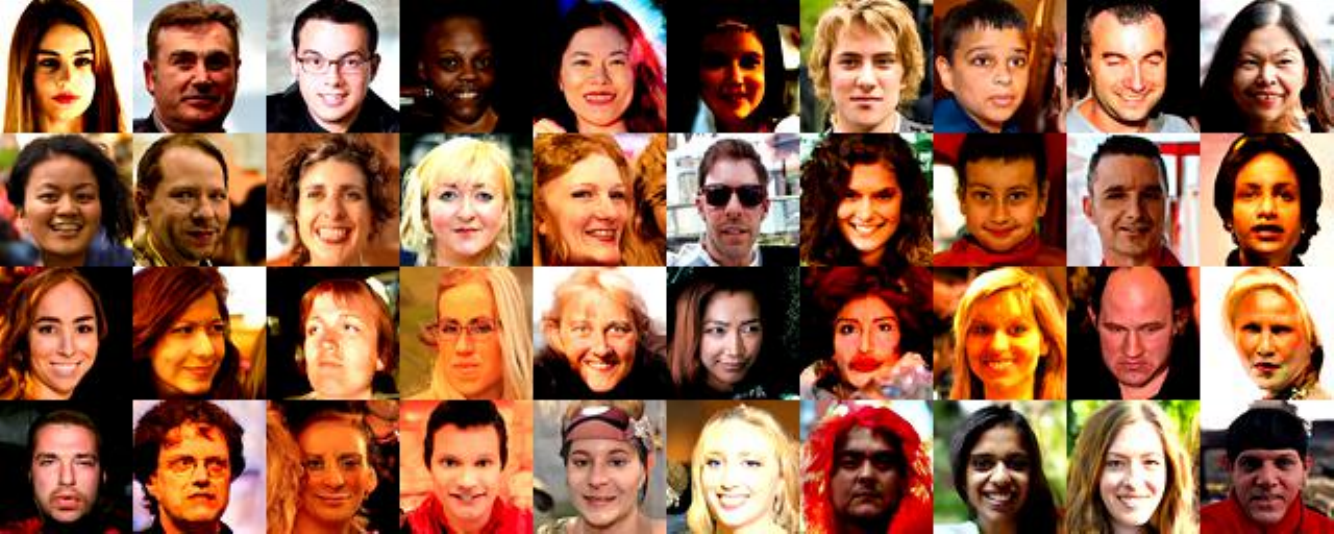}
      \caption{DPM-Solver-2. NFE=\(\nfeb\)}
  \end{subfigure}
  \begin{subfigure}[b]{0.48\linewidth}
      \includegraphics[width=\linewidth]{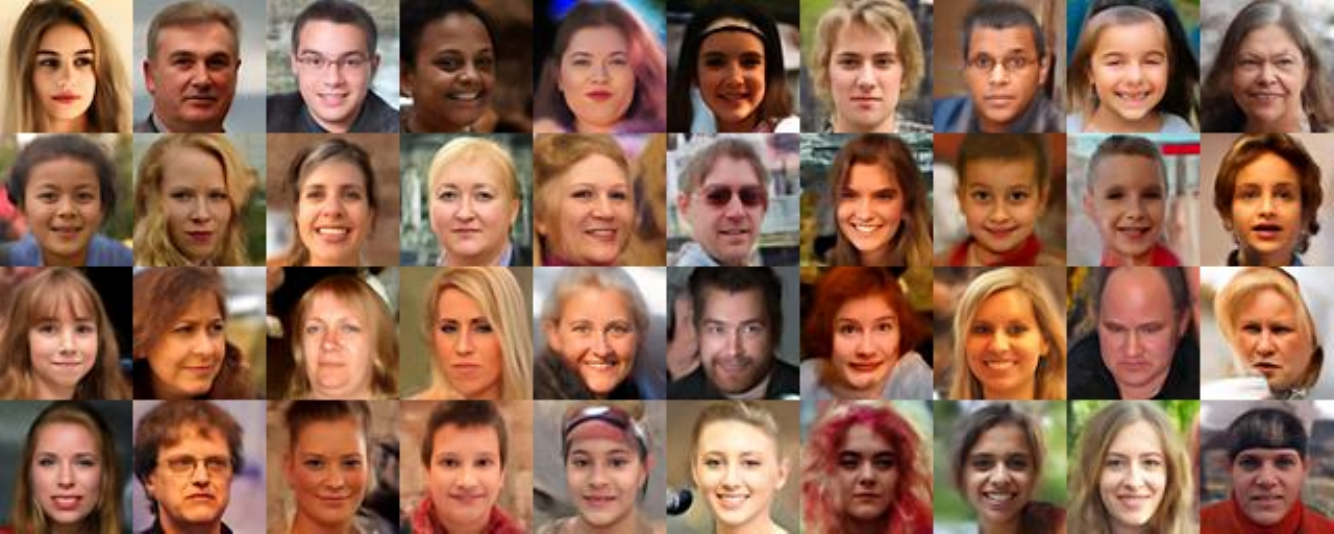}
      \caption{\(\ours\). Para. NFE=\(\nfea\)}
  \end{subfigure}
  \begin{subfigure}[b]{0.48\linewidth}
      \includegraphics[width=\linewidth]{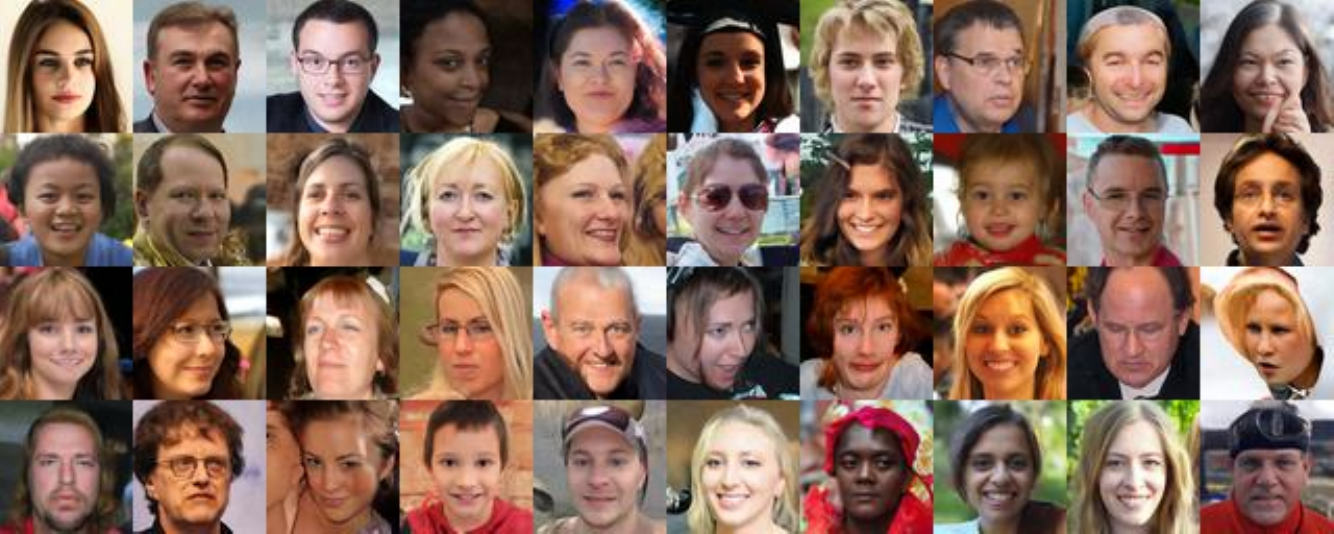}
      \caption{\(\ours\). Para. NFE=\(\nfeb\)}
  \end{subfigure}
  \caption{Qualitative result on FFHQ 64$\times$64 (\(\nfea\) and \(\nfeb\) NFEs)}
  \label{fig:sup_grid_ffhq_3}
\end{figure*}

\begin{figure*}[t]
  \centering
  \begin{subfigure}[b]{0.48\linewidth}
      \includegraphics[width=\linewidth]{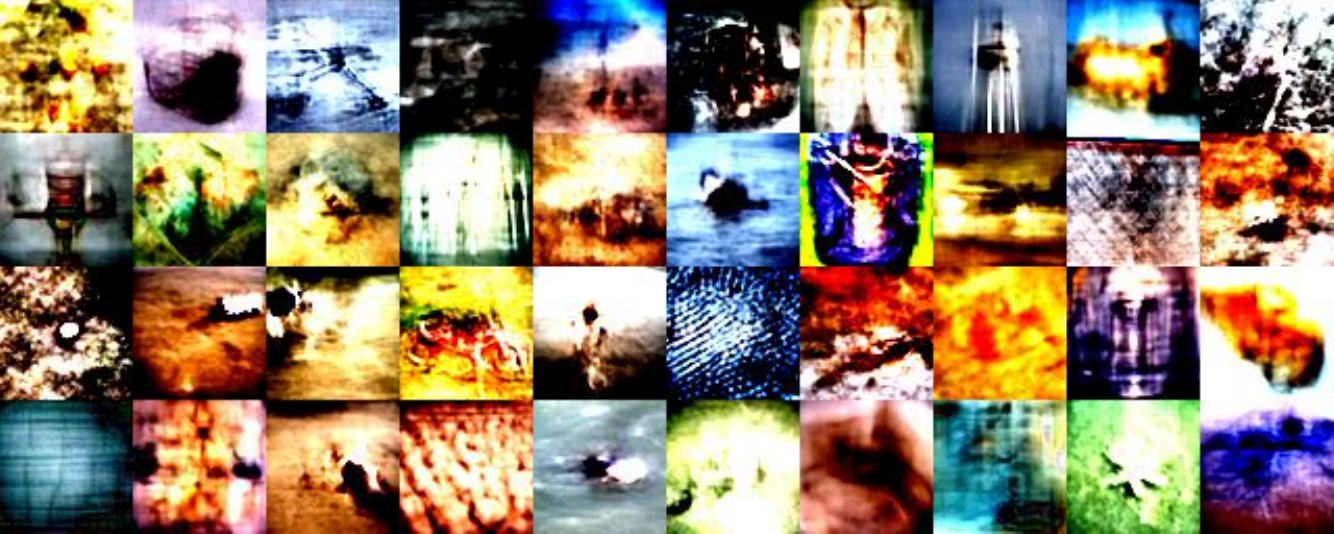}
      \caption{DPM-Solver-2. NFE=\(\nfea\)}
  \end{subfigure}
  \begin{subfigure}[b]{0.48\linewidth}
      \includegraphics[width=\linewidth]{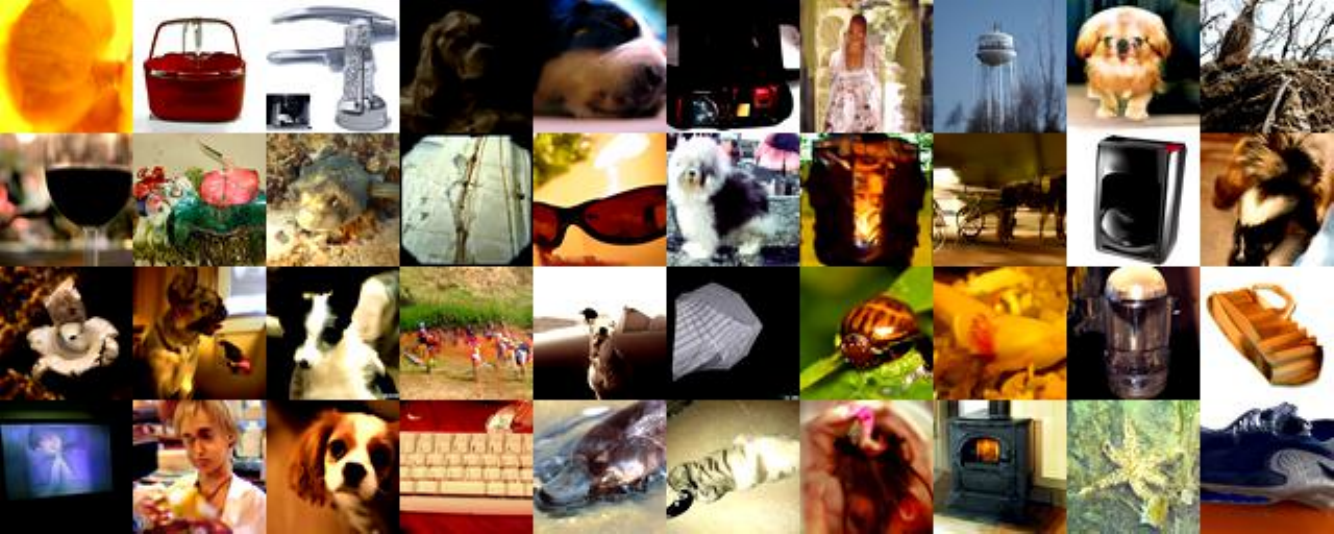}
      \caption{DPM-Solver-2. NFE=\(\nfeb\)}
  \end{subfigure}
  \begin{subfigure}[b]{0.48\linewidth}
      \includegraphics[width=\linewidth]{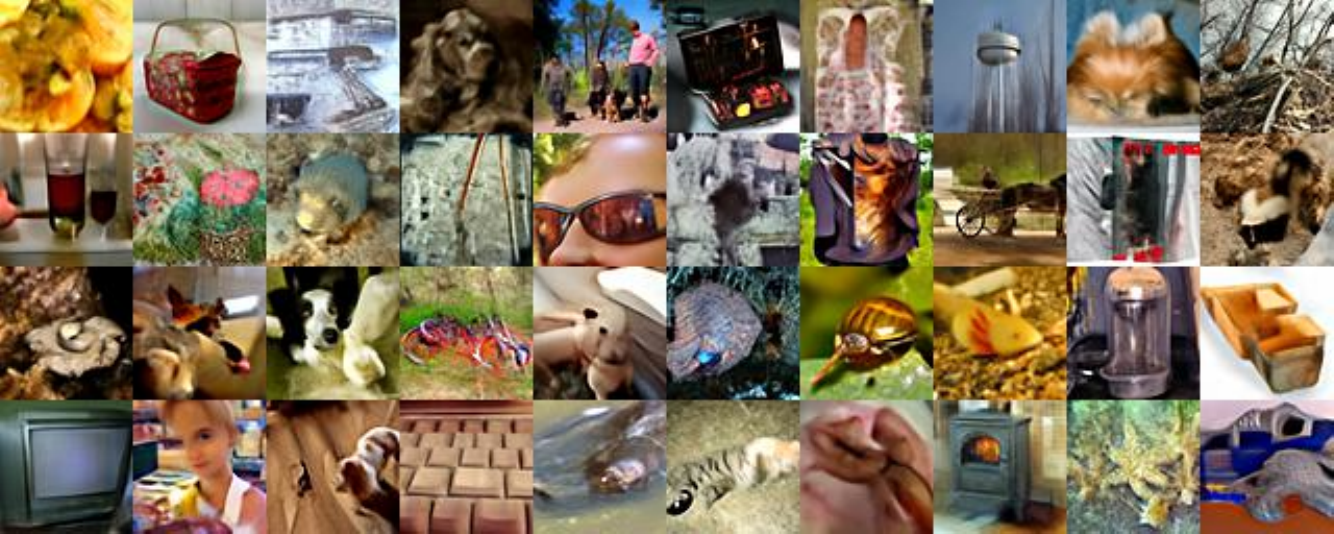}
      \caption{\(\ours\). Para. NFE=\(\nfea\)}
  \end{subfigure}
  \begin{subfigure}[b]{0.48\linewidth}
      \includegraphics[width=\linewidth]{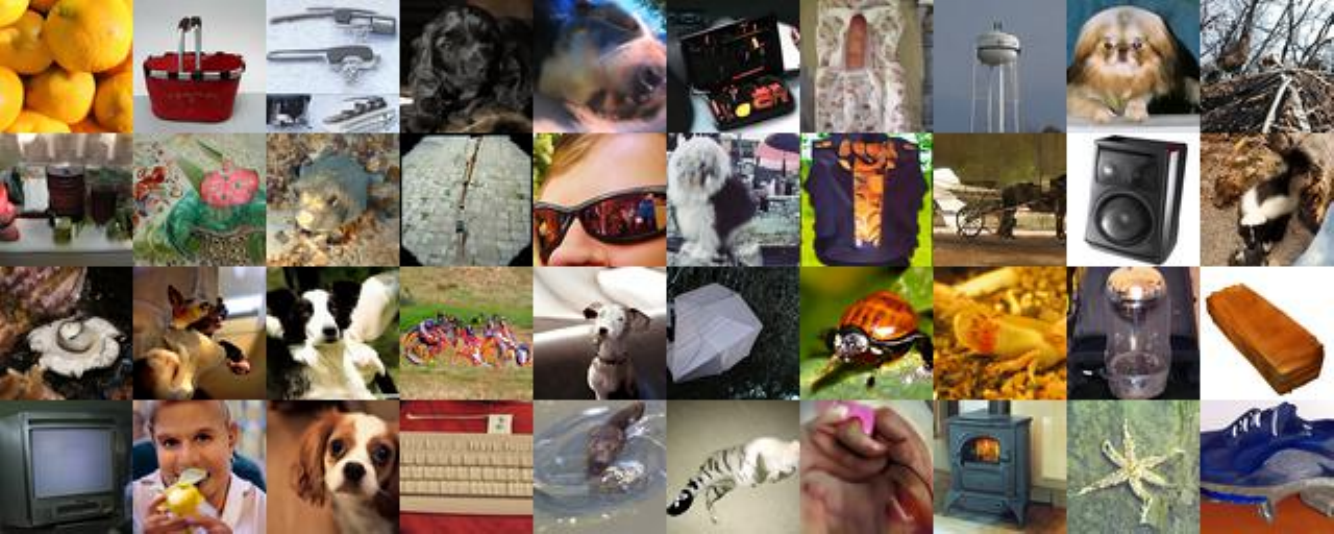}
      \caption{\(\ours\). Para. NFE=\(\nfeb\)}
  \end{subfigure}
  \caption{Qualitative result on ImageNet 64$\times$64 (\(\nfea\) and \(\nfeb\) NFEs)}
  \label{fig:sup_grid_img_3}
\end{figure*}

\end{document}